\documentclass[final,5p,times,twocolumn]{elsarticle}

\usepackage{amssymb}

\usepackage{lineno}

\usepackage{algorithmic}
\usepackage{algorithm}
\usepackage{multirow}
\usepackage{multicol}
\usepackage{mathtools}
\usepackage{graphicx}
\usepackage{caption,setspace}
\usepackage{array}
\usepackage{booktabs}
\usepackage{url}
\usepackage[colorinlistoftodos]{todonotes}

\usepackage{hyperref}
\hypersetup{%
	pdfborder = {0 0 0}
}

\DeclareMathOperator*{\argmin}{arg\,min}
\usepackage{bm}
\usepackage{subfigure}

\usepackage{algorithm}
\usepackage{algorithmic}
\newcommand{\algorithmicbreak}{\textbf{break}}
\newcommand{\BREAK}{\STATE \algorithmicbreak}

\usepackage{enumitem}

\journal{Elsevier}

\begin{document}

@article
\{YIN2023104641,\\
title = {Semantic localization on BIM-generated maps using a 3D LiDAR sensor},\\
journal = {Automation in Construction},\\
volume = {146},\\
pages = {104641},\\
year = {2023},\\
issn = {0926-5805},\\
doi = {https://doi.org/10.1016/j.autcon.2022.104641},\\
url = {https://www.sciencedirect.com/science/article/pii/S0926580522005118},\\
author = {Huan Yin and Zhiyi Lin and Justin K.W. Yeoh},\\
keywords = {Building information modeling, LiDAR, Localization, Semantic, Iterative closest point}\\
\}

\newpage


\begin{frontmatter}



\title{Semantic localization on BIM-generated maps using a 3D LiDAR sensor}

\author[address1,address2]{Huan Yin\corref{cor1}}
\author[address3]{Zhiyi Lin}
\author[address2]{Justin K.W. Yeoh\corref{cor1}}

\cortext[cor1]{Corresponding author}

\address[address1]{Department of Electronic and Computer Engineering, Hong Kong University of Science and Technology, Hong Kong SAR}
\address[address2]{Department of Civil and Environmental Engineering, National University of Singapore, Singapore}
\address[address3]{China Mobile (Zhejiang) Innovation Research Institute Co., Ltd, Hangzhou, China}

\begin{abstract}	
Conventional sensor-based localization relies on high-precision maps, which are generally built using specialized mapping techniques involving high labor and computational costs. In the architectural, engineering and construction industry, Building Information Models (BIM) are available and can provide informative descriptions of environments. This paper explores an effective way to localize a mobile 3D LiDAR sensor on BIM-generated maps considering both geometric and semantic properties. First, original BIM elements are converted to semantically augmented point cloud maps using categories and locations. After that, a coarse-to-fine semantic localization is performed to align laser points to the map based on iterative closest point registration. The experimental results show that the semantic localization can track the pose successfully with only one LiDAR sensor, thus demonstrating the feasibility of the proposed mapping-free localization framework. The results also show that using semantic information can help reduce localization errors on BIM-generated maps.
\end{abstract}


\end{frontmatter}


\section{Introduction}
\label{intro}

Localization is an essential capability for robot navigation that estimates the position and orientation of a robot. Almost all construction robots, whether tele-operated or autonomous require the estimated poses from the localization module to achieve safe human operation or self-navigation \cite{peel2018localisation,kim2018slam}. 

With the development of sensor technologies, indoor localization can be achieved by deploying AprilTag \cite{kayhani2022tag}, ultra-wideband \cite{prorok2014accurate} or other signal emitters in buildings. Such methods rely on the distribution of sensors and inherently lack flexibility within large built-up environments. Instead of deploying sensors in such environments, a more popular approach is to utilize the perception capabilities of onboard sensors, such as laser scanners and cameras, which can improve the generalizability of the localization module in large scenes.

In robotics, a general localization approach is the Simultaneous Localization and Mapping (SLAM) system \cite{zhang2014loam,qin2018vins}, which achieves mapping and localization simultaneously using onboard sensors. However, for some long-term applications that operate under stable conditions, i.e., a quadruped robot working daily on building inspection, the mapping process of SLAM is redundant because the generated map is almost invariant in each run of SLAM. Besides that, a complete SLAM system requires high computing resources and multiple additional modules to guarantee both efficiency and accuracy, such as online loop closing \cite{yin2018locnet}, map management \cite{burki2018map} and sensor calibration \cite{liu2022targetless}, leading to high costs for long-term operations.

A two-stage approach is widely used to address this problem: first mapping and then metric localization within the known map \cite{krusi2015lighting,ding2019persistent}. In this approach, map building is required only once and after that localization on the map is able to handle the pose tracking for long-term operations, thereby reducing the complexity of repetitive SLAM processes. In the Architectural, Engineering and Construction (AEC) industry, some models or representations are directly available, such as Computer-aided Design (CAD) or Building Information Models (BIM). These map-like representations contain informative measurements that are human readable. We propose the idea that high-cost pre-mapping may not always be necessary in known built environments, and mapping-free localization could be an alternative choice.

On the other hand, architectural CAD and BIM are designed for construction and building management so they are not localization-oriented. To bridge the gap between architectural models and pose estimation, a number of research works proposed to align laser points or visual images to the as-designed models \cite{kim2013fully,asadi2019real,blum2021precise,chen2022align}. However, almost all alignment approaches were performed using only geometric properties of observed points and models. In recent years, with the popularity of BIM, semantically rich models provide high-level semantic information for building construction and management. This semantic information is helpful for scene understanding and is easy to obtain compared to traditional CAD models. Thus, we hypothesize that \emph{the semantic property of BIM could help improve the performance of robot localization}.

Deep learning techniques have been widely used to build a semantic localization method for feature extraction and data association \cite{chen2019suma++,wei2019vision,acharya2022single}. Large amounts of labeled data is required to train neural networks and these existing works can not guarantee the generalization ability in unseen environments. In addition, learning-based localization methods are generally computationally expensive with high time costs in the inference stage, especially for real-time 3D LiDAR points, leading to inefficient applications when using resource-constrained devices.

It is concluded that a desirable localization in BIM requires both effectiveness and efficiency for applications in the real world. In this paper, a novel learning-free framework is proposed to achieve localization on BIM-generated maps with only one 3D mobile LiDAR sensor, as shown in Figure~\ref{framework}. Specifically, the entire framework consists of two pipelines: offline BIM-to-Map conversion and online coarse-to-fine localization. The offline pipeline can convert the as-designed BIM to semantically augmented point cloud maps. After that, these semantic maps are utilized to filter input laser points, and pose tracking is achieved by performing Iterative Closest Point (ICP) on the filtered points. The entire framework requires no deep learning for feature extraction or pose regression, making it totally interpretable. Finally, extensive experiments are conducted using our self-collected multi-session dataset in a real-world university building.

Our major contributions are summarized as follows:
\begin{itemize}
\item A pipeline is built to effectively convert BIM to semantic point cloud maps, which can bridge the gap between digital representations and localization-oriented maps. The pipeline does not require manually labeled data.
\item The semantic information of BIM is utilized to filter laser points and weight data associations, thus building a semantic-aided LiDAR localization on BIM-generated maps.
\item The proposed method is validated in a real-world building via multi-session tests. Experimental results validate the feasibility and effectiveness of semantic localization on BIM-generated maps. 
\end{itemize} 

The rest of this paper is organized as follows: the related work is presented in Section~\ref{related}. The proposed semantic localization framework is introduced in Section~\ref{method}. Section~\ref{exp} reports the experimental set-up and results on our self-collected datasets. Section~\ref{conclusion} presents conclusions and future studies.

\section{Related Work}
\label{related}

\begin{table*}[t]
\captionsetup{justification=centering}
\renewcommand\arraystretch{1.5}
\begin{center}
	\caption{Selected LiDAR-based localization on CAD or BIM-based maps}
	\label{studies}
	\begin{tabular}{p{1cm}<{\centering}|m{2cm}<{\centering}m{4.0cm}<{\centering}m{7cm}<{\centering}m{2cm}<{\centering}}
		\hline
		\hline
		Ref. & Source of map & Sensors for localization & Task / Method / Highlight &  Experimental validation \\
		\hline
		\cite{boniardi2017robust} & floor plan & 2D LiDAR & robot localization via pose-graph SLAM and GICP-based scan-to-map matching  & real world \\
		\cite{boniardi2019pose} & floor plan & 2D LiDAR & increase the robusteness and efficiency of \cite{boniardi2017robust} & real world  \\
		\cite{wang2019glfp} & floor plan & 3D LiDAR and wheel encoder and gyroscope & global localization using edge features and pose tracking using factor graph  & real world \\
		\cite{blum2021precise} & 3D floor plan & 3D LiDAR and cameras & selective ICP-based localization by integrating semantic information from images & real world (stationary) \\
		\cite{gao2022fp} & floor plan & 3D LiDAR and IMU & build a novel nearest neighbour field on CAD for efficient feature registration & real world \\
		\hline
		\cite{follini2020bim} & BIM & 2D LiDAR and IMU (Husky A200 MRP) & propose BIM-generated time-dependent maps, and use AMCL \cite{amcl} for localization & Gazebo and real world \\
		\cite{kim2021development} & BIM & 2D LiDAR and odometry (Neobotix MMO-500)  & develop a robotic wall painting system, including AMCL for localization & Gazebo  \\
		\cite{kim2022bim} & BIM & 2D LiDAR and odometry (Turtlebot2) & create a semantic building world for task planning, including AMCL for localization & Gazebo \\
		\cite{hendrikx2021connecting} & BIM & 2D LiDAR and wheel odometry  & robot localization via feature matching in spatial-semantic database and factor graph & real world  \\
		Ours & BIM & 3D LiDAR  & semantic localization via filtering points using BIM and performing semantic ICP & real world  \\
		\hline
		\hline
	\end{tabular}
\end{center}
\end{table*}

\subsection{Robot localization on CAD or BIM-based maps}
\label{bimbased}

Many research publications have reviewed pose estimation topics from different perspectives, including deep learning-based \cite{chen2020survey}, sensor-based \cite{lowry2015visual,elhousni2020survey,yin2021rall}, etc. These research papers mainly focused on robot localization on visual or lidar maps, which require SLAM or data collection for pre-mapping. In this study, we propose to achieve mapping-free localization, and the related works mainly focus on CAD or BIM-based localization in this subsection.

Floor plans or point clouds can be generated from CAD models for LiDAR localization \cite{li2020online,boniardi2017robust,boniardi2019pose,wang2019glfp,blum2021precise,gao2022fp}. Researchers in \cite{li2020online} proposed to localize a 2D laser scanner on floor plans and hand-drawn maps using stochastic gradient descent. At the back-end, pose graphs were built in \cite{boniardi2017robust,boniardi2019pose} to increase the localization robustness on floor plan-based maps. As for localization in 3D space, ICP-based alignment is considered an effective method to track the robot pose \cite{blum2021precise}. Other than the point-based ICP method, meshes were also used for robot global localization without the need of an initial guess in \cite{dreher2021global}. Recently, researchers in \cite{ercan2020online} proposed a novel interface to connect building construction and map representation, which could also detect deviations between as-designed and as-built models via localization results.

Compared to traditional CAD models, BIM is more interoperable in the construction industry and contains more semantic information that may be suitable for robot navigation. For single-frame-based localization, photogrammetric point clouds can be aligned to BIM \cite{chen2022align} for camera pose estimation from scratch. As for pose tracking, visual-based pose tracking was also demonstrated to be effective \cite{asadi2019real}, in which camera poses were estimated by aligning images to BIM models. In \cite{wei2019vision}, learning-based visual localization was proposed for facility operations and management. Generally, deep learning-based methods rely on pre-trained neural networks for feature extraction or pose regression, bringing difficulties for debugging and deployment in the real world. Researchers in \cite{hendrikx2021connecting} extracted semantic features without learning and also performed robot localization in BIM using 2D laser scans. Some recent studies \cite{follini2020bim,kim2021development,kim2022bim} also used BIMs as maps in Gazebo for robot planning tasks, in which Adaptive Monte Carlo Localization (AMCL) \cite{amcl} was used to track the mobile robot pose. These recent studies inspire us that it is feasible to integrate BIM into robotic systems as maps.

\begin{figure*}[t]
\centering
\includegraphics[width=\linewidth]{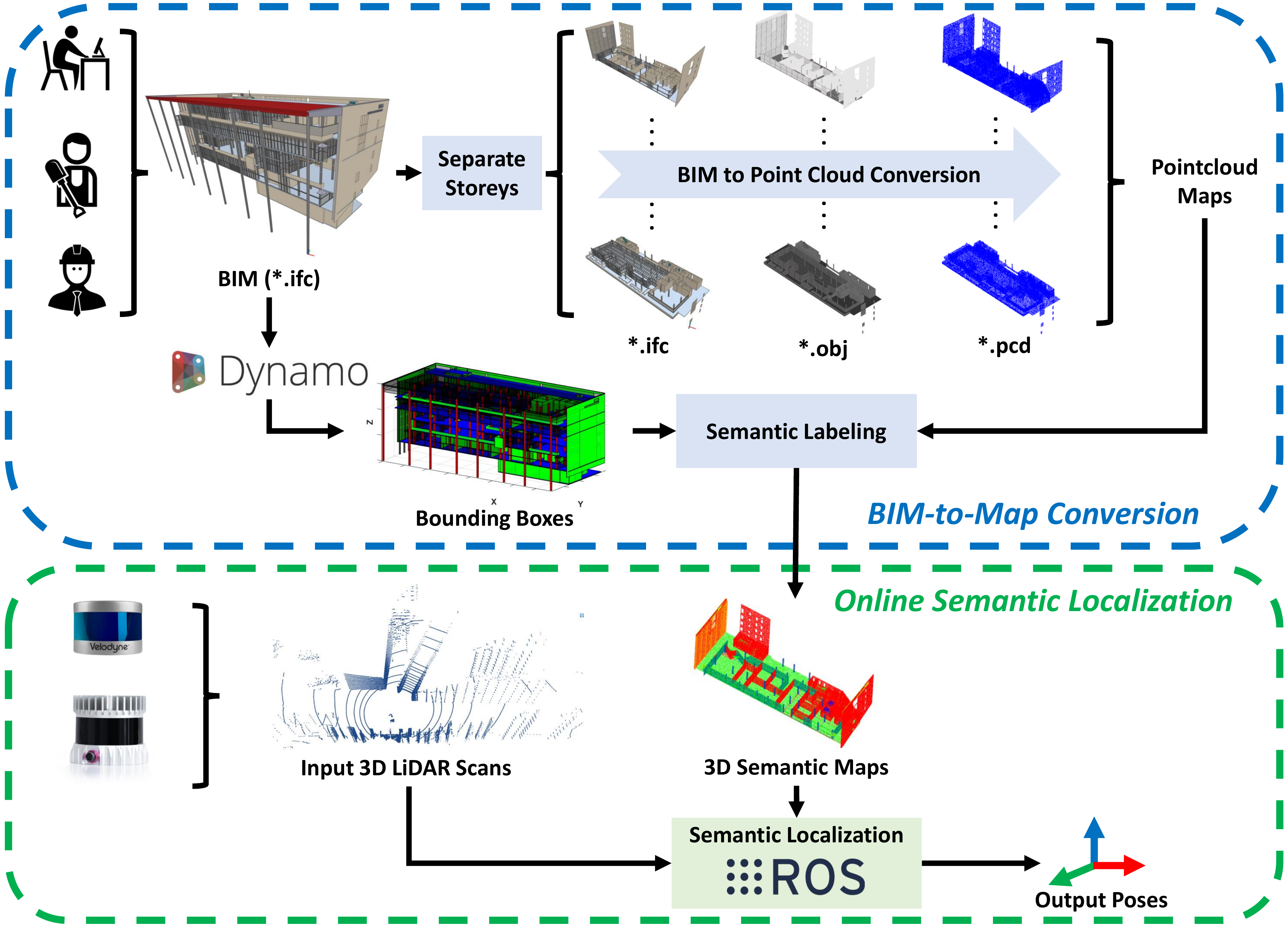}
\caption{Our proposed mapping-free localization framework consists of two pipelines: offline BIM-to-Map conversion and online semantic localization. 
}
\label{framework}
\end{figure*}

Table~\ref{studies} presents several representative studies with brief descriptions. Most of these works in the table used some other sensors as support, such as wheel odometry \cite{wang2019glfp,hendrikx2021connecting} and IMU \cite{gao2022fp,follini2020bim}, which will increase the robustness of the localization module. But on the other hand, these sensors bring higher hardware costs and potential calibration problems. Besides that, many works utilized state estimators at the back-end, i.e., graph optimization-based \cite{boniardi2019pose,wang2019glfp} and particle filter-based \cite{follini2020bim,kim2021development}, which are mainly designed for multiple sensor fusion. These methods are more computational expensive compared to those that only use scan matching at the front-end \cite{blum2021precise}. In this study, only one LiDAR scanner is used for pose estimation, and the method can be deployed on a resource-constrained laptop device.

Furthermore, most studies only used the geometric information for localization. In \cite{blum2021precise} and \cite{hendrikx2021connecting}, semantic information was integrated into LiDAR localization system. Specifically, in \cite{blum2021precise}, semantic information was generated from learning-based image segmentation, and the researchers validated the method using a stationary robot. In \cite{hendrikx2021connecting}, semantic features were extracted from laser scans and then matched to BIM-based database. The features were sparse compared to laser points, and a factor graph was also built to achieve pose estimation in \cite{hendrikx2021connecting}. Overall, our method is inspired by these existing studies, and we propose to build a semantic-aided LiDAR-only localization on BIM-generated point cloud maps.

\subsection{Semantic mapping and localization}
\label{relatedsemantic}

Semantic mapping and localization is a popular topic in the robotics community. Compared to geometric-only localization, semantic localization is able to closely mimic human understanding of the real world.

Semantic information is easy to extract from visual images. A typical semantic-based visual localization is retrieving query images from database, namely visual place recognition or global localization \cite{lowry2015visual}. Semantic information is also helpful for metric pose estimation \cite{toft2018semantic}. Almost all semantic-based visual localization require deep neural networks for feature extraction at the front-end.

As for point cloud-based localization, researchers also proposed to use semantics to enhance the data matching. A semantic ICP-based registration was proposed and validated in RGBD dataset \cite{parkison2018semantic}. Similarly, semantic ICP was also used in \cite{chen2019suma++} to localize a vehicle on the road. In \cite{chen2020sloam}, semantic-based LiDAR SLAM was tested in challenging forest environments, where tree trunks can be segmented by neural networks. Overall, semantic information was obtained by manually labeled data and trained networks in \cite{parkison2018semantic,chen2019suma++,chen2020sloam}. In this study, the input laser points are labeled using the BIM-generated maps, which could make the localization module more efficient.

\section{Methodology}
\label{method}
\subsection{Overview}

\label{overview} 

Given a mobile LiDAR scanner and a BIM file, we denote the input LiDAR data as $\mathcal{P}_k$ at timestamp $k$ and the global point cloud map as $\mathcal{M}$. The timestamp index is omitted for simplified representation of a single time instance in this paper. The main problem of metric localization is how to align $\mathcal{P}$ to the reference $\mathcal{M}$ by estimating a transformation $\mathbf{T} = \left[ \mathbf{R}, \mathbf{t}\right]\in\text{SE(3)}, \mathbb{R}^{4\times4}$, where $\mathbf{R}$ and $\mathbf{t}$ are estimated rotation and translation respectively. The alignment must be precise and efficient to guarantee the estimation of $\mathbf{T}_{k=1,2,\cdots} $with sequential inputs  $\mathcal{P}_{k=1,2,\cdots}$, or namely pose tracking. In the context of this paper, pose tracking and localization are deemed to have the same meaning. 

As shown in Figure~\ref{framework}, the proposed semantic localization framework consists of two pipelines: offline BIM-to-Map conversion and online semantic localization with a mobile LiDAR scanner. The offline pipeline converts original BIM file to a localization-oriented point cloud map $\mathcal{M}$, and also labels the map points with categories from BIM. The semantic localization pipeline is designed to track the mobile LiDAR scanner based on the reference $\mathcal{M}$ and inputs $\mathcal{P}_{k=1,2,\cdots}$ .

\subsection{From BIM to semantic maps}
\label{bimtomap}

Within the AEC industry, BIMs can be created by many software tools and has been used to support various construction processes, such as building inspection \cite{tan2021automatic} and quality management \cite{ma2018construction}. To achieve robot or sensor localization in Euclidean space, precise metric maps are required instead of modeled information. In this study, the first challenge is how to generate localization-oriented point cloud maps from BIM files.

The BIM-to-Map conversion consists of three steps, shown in the upper part of Figure~\ref{framework}. The whole BIM of one building is first split into several separate BIMs according to different storeys. After that, the digital BIM files are converted to obj files using IfcOpenShell \cite{IfcOpenShell}. Finally, 3D point clouds are sampled from triangular meshes of obj files with a defined density \cite{cignoni1998metro}. There are several other sampling strategies in some software \cite{MeshLab,CloudCompare}, such as Monte-Carlo Sampling. Considering that density value is easily understood and defined by most users, we decide to use this strategy for point cloud generation in this paper. The final point cloud maps can be regarded as sub-maps of each floor in the building.

Our experience has shown that it is better not to change the sequence of this conversion. In other words, if geometries are extracted from the whole BIM first without separating into storeys, the storey information of BIM is not captured. It then becomes more challenging to split a large geometric model or point cloud map into storey-based sub-maps. 

\begin{figure}[t]
\centering
\subfigure[BIM for map generation]{
	\includegraphics[width=8.75cm]{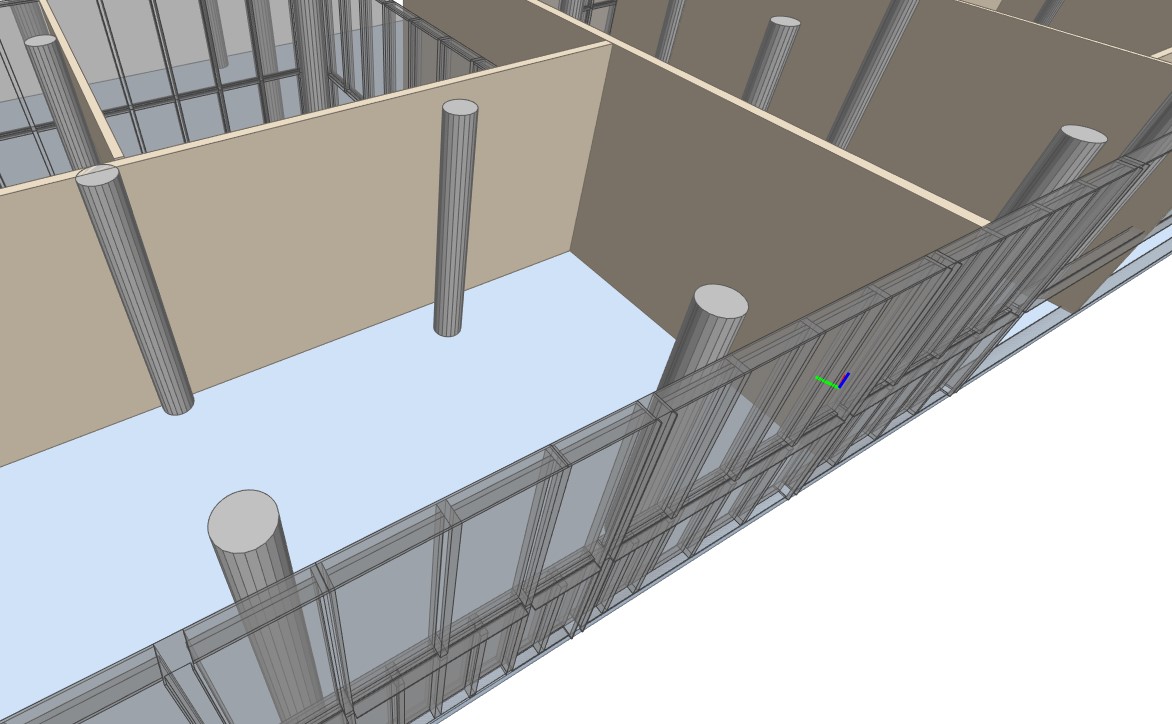}}
\subfigure[Map points in bounding boxes]{
	\includegraphics[width=8.75cm]{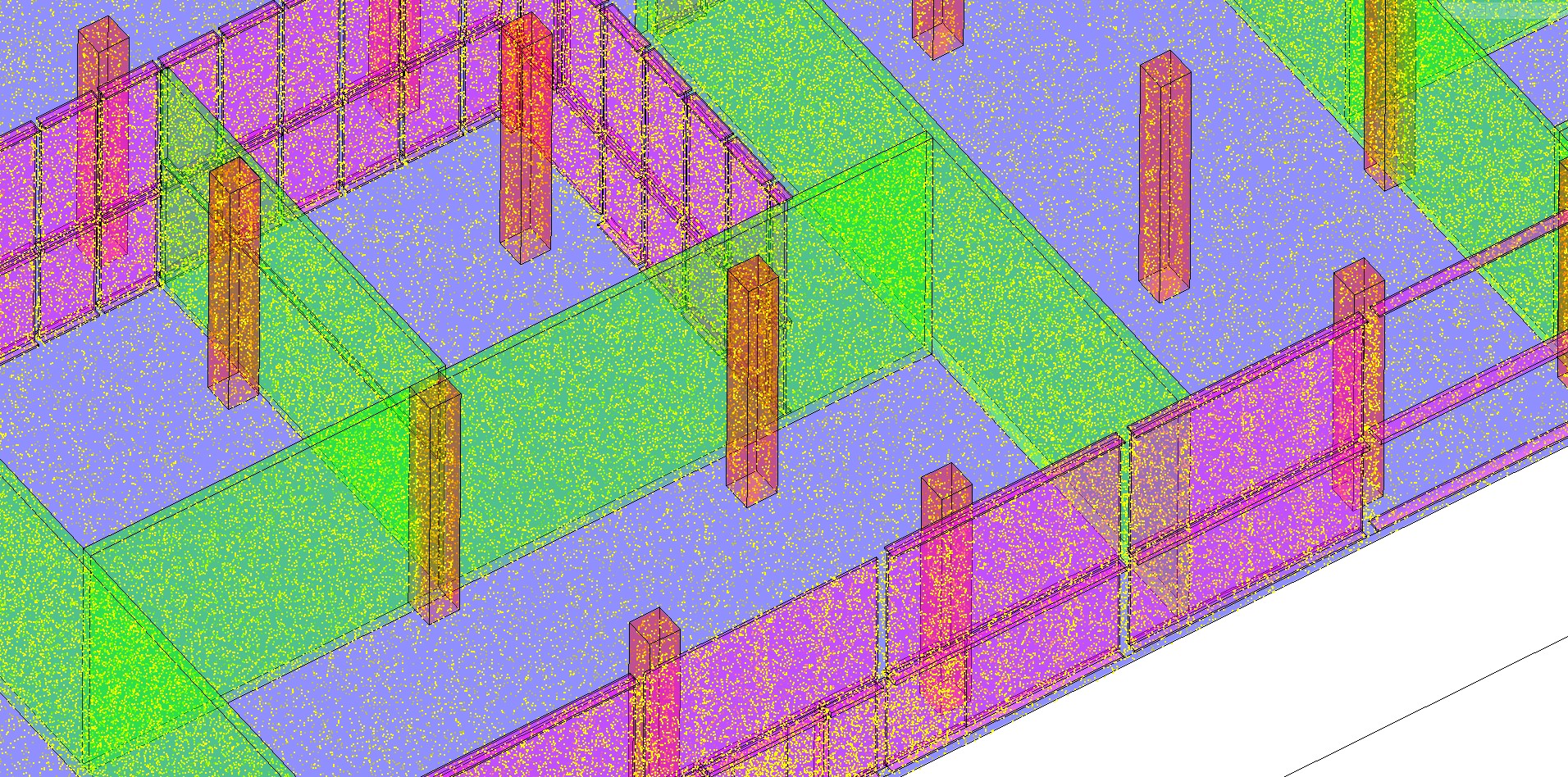}} 
\caption{BIM, bounding boxes and map points at the same place. Box colors stand for different categories: blue for ``Floor'', red for ``Column'', green for ``Wall'' and pink for ``Window'' respectively. Yellow points in boxes are map points generated from BIM. }
\label{box}
\end{figure}

Until this step, the point cloud maps are generated using the geometric and storey information of BIM. However, BIM also contains rich semantic information compared to a typical geometry model and these semantic properties can be updated manually and dynamically, which may bring potential advantages for robot navigation. Thus we also propose to integrate the semantics into the offline BIM-to-Map process, thus maximizing the utilization of information from BIM for metric localization in this paper.

\begin{algorithm}[t]
\caption{Semantic Labeling of Map Points Using BIM}
\label{labeling}
\begin{algorithmic}
	\REQUIRE ~~\\ 
	A point cloud map from BIM: $\mathcal{M}= \left\{ \mathbf{m}_i\right\}$ \\
	Bounding boxes from BIM: $\mathcal{D}=\{\mathbf{d}_j\}$   \\
	Category labels of each box: $\mathcal{C}=\{\mathbf{c}_j\}$ 
	\ENSURE 
	\STATE Build a k-d tree on $\mathcal{D}$
	\FOR{$\mathbf{m}_i$ in $\mathcal{M}$}
	\STATE $\mathcal{D}_\text{nn}$ $\leftarrow$ nearest neighbor search of $\mathbf{m}_i$ in the tree
	\FOR{$\textbf{d}_j$ in $\mathcal{D}_\text{nn}$}
	\IF{Creteria~(\ref{criteria1})}
	\STATE $c\left(\mathbf{m}_i\right) = \mathbf{c}_j$
	\BREAK 
	\ENDIF
	\ENDFOR
	\ENDFOR
\end{algorithmic}
\end{algorithm}

To achieve this, an automated approach for map labeling is used, which is simple but effective compared to the manually labeling process \cite{chen2020sloam}. Let $\mathbf{m}_i$ be a map point of $\mathcal{M}$. To label the map point $\mathbf{m}_i$, Dynamo \cite{Dynamo} is used to extract the category labels $\mathcal{C}$ and bounding boxes $\mathcal{D}$ of all elements in BIM. The minimum and maximum location points of one bounding box $\mathbf{d}_j$ are notated as $\mathbf{d}_{\min}=\left[x_{\min}, y_{\min}, z_{\min}\right]^\top$ and $\mathbf{d}_{\max}=\left[x_{\max}, y_{\max}, z_{\max}\right]^\top$, which can represent the coverage of $\mathbf{d}_j$ in 3D space. With the extracted bounding boxes and labels, we then retrieve all the boxes and classify whether $\mathbf{m}_i=\left[x_i,y_i,z_i\right]^\top$ is in a specific box. The classification criteria is as follows:

\begin{equation}\label{criteria1}
\mathbf{m}_i\;\text{is in}\;\mathbf{d}_j = \left\{\begin{array}{cc}
	& x_{\min} \leq x_i \leq x_{\max}\\
	1\ , & y_{\min} \leq y_i \leq y_{\max} \\
	& z_{\min} \leq z_i \leq z_{\max} \\
	\\
	0\ , & \text{else}.
\end{array}
\right.
\end{equation}

Specifically, to accelerate the semantic labeling process, a K-Dimensional (k-d) tree is built based on the center points of $\mathcal{D}$. In summary, the proposed semantic labeling process is presented in detail in Algorithm~\ref{labeling}. Note that the labels of $\mathcal{C}$ are not unique, which means some map points are in different boxes but with the same category label $c(\cdot)$, e.g., columns are with the same category label ``Column''. An example is presented in Figure~\ref{box} to help better understand the semantic mapping process, in which different colors represent different categories.

\par One might argue that the bounding box extraction in Dynamo is not so accurate and some map points could exist in multiple boxes, i.e, points may lie on the boundary of columns and floors, leading to the ambiguity of semantic map building. These are termed as mixed labeled points. In reality, there are relatively few of such ambiguously labeled points on the boundary of multiple elements. 

Another problem is that the bounding box is not oriented from Dynamo in this study, and it might be so large that could cover other elements. For example, a thin wall is from (0,0,0) to (10,10,10), but the size of its box is 10$\times$10$\times$10, which will make some points incorrectly labeled in this large box. Figure~\ref{wrong_box} presents two cases in the following experimental section. Overall, these mixed labeled or incorrectly labeled points have impact on the localization performance, but will not cause localization failures. This will be validated in the experimental section.

\subsection{Semantic localization on BIM-generated maps}
\label{semicp}

With the generated semantic map $\mathcal{M}$, the online semantic localization pipeline aims to estimate transformations $\mathbf{T}_{k=1,2,\cdots}$ with inputs $\mathcal{P}_{k=1,2,\cdots}$. At timestamp $k$, the kernel of the localization problem is to align a LiDAR scan $\mathcal{P}$  to $\mathcal{M}$, which can be achieved by minimizing the error function $e\left(\cdot\right)$ between two point clouds, stated as follows:

\begin{equation} \label{error1}
\mathbf{T} = \argmin_{\mathbf{T}\in\text{SE(3)}} \left( e\left(\mathcal{M}, \mathbf{T}\mathcal{P}\right) \right)
\end{equation}

Then, data association is required to build the error function. We denote the data association as $\mathcal{A}=\{ \left(\mathbf{p}, \mathbf{m}\right); \mathbf{p} \in \mathbf{T}\mathcal{P}, \mathbf{m} \in \mathcal{M}\}$, where $\left(\mathbf{p}, \mathbf{m}\right)$ is a match between the transformed input LiDAR scan and the reference map. The error function is formulated as follows:

\begin{equation} \label{error2}
\mathbf{T} = \argmin_{\mathbf{T}\in\text{SE(3)}} \left( \sum^\mathcal{A} e\left(\mathbf{p}, \mathbf{m}\right) \right)
\end{equation}

Furthermore, to build a robust data association, some relations can be used to build weights $\mathcal{W}=\{ w\left(\mathbf{p}, \mathbf{m}\right) \in \left[0,1\right]; \left(\mathbf{p}, \mathbf{m}\right) \in \mathcal{A}\}$. $\mathcal{W}=\mathbf{1}$ means all point matches are used without weights in error minimization. Consequently, the error function is as follows:

\begin{equation} \label{error3}
\mathbf{T} = \argmin_{\mathbf{T}\in\text{SE(3)}} \left( \sum^\mathcal{A} w\left(\mathbf{p}, \mathbf{m}\right)e\left(\mathbf{p}, \mathbf{m}\right) \right)
\end{equation}

\begin{figure}[t]
\centering
\includegraphics[width=9cm]{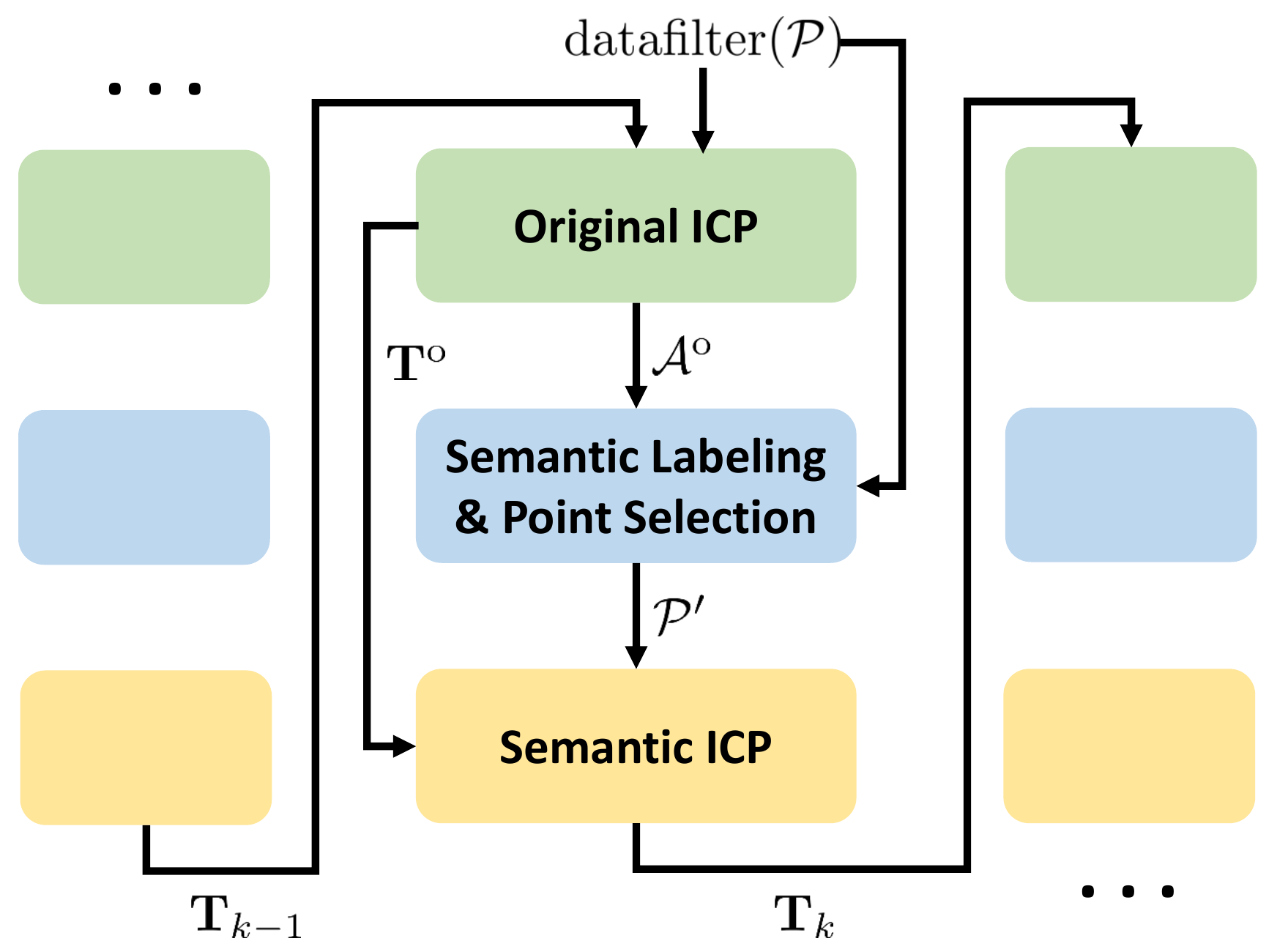}
\caption{Proposed three-step semantic localization pipeline.}
\label{semanticloc}
\end{figure}

Various point cloud registration methods have been proposed to minimize the error function in Equation~(\ref{error3}). Generally, among these methods, ICP is one of the most widely used methods in the robotics community \cite{besl1992method,pomerleau2015review}. Equation~(\ref{error3}) is solved via ICP iteratively and the matches $\mathcal{A}$ and weights $\mathcal{W}$ are updated in each iteration.

ICP has many variants in different robotic or computer vision applications. For the error term $e\left(\cdot\right)$, we estimate the normal vectors $\mathbf{n}$ of each map point $\mathbf{m}$ and use point-to-plane ICP for pose estimation. Thus the error metric term  in Equation~(\ref{error3}) can be expressed as:

\begin{equation} \label{error4}
e\left(\mathbf{p}, \mathbf{m}\right)=\lVert \left(\mathbf{R}\mathbf{p} + \mathbf{t} - \mathbf{m} \right) \cdot \mathbf{n} \rVert_{2}
\end{equation}
where $\mathbf{R}$ and $\mathbf{t}$ are rotation and translation of $\mathbf{T}$ respectively.

As for the weight term $w\left(\cdot\right)$ in Equation~(\ref{error3}), we intend to use semantic associations to weight the data associations in this paper. Generally, the semantic associations are built from semantically labeled maps and sensor readings, as presented in previous studies \cite{parkison2018semantic,chen2019suma++}. In this study, semantic maps $\mathcal{M}$ can be built from BIM, but the raw input scan $\mathcal{P}$ are not labeled, leading to a difficulty in building semantic associations. Thus, the challenge is how to label the input laser points effectively and efficiently on BIM-generated maps. Then the labeled laser points can be utilized for a semantic-aided localization on semantic maps.

\begin{algorithm}[t]
\caption{Three-step Semantic Localization} 
\label{semanticicp}
\begin{algorithmic}
	\REQUIRE ~~\\ 
	Semantic map from BIM using Algorithm~\ref{labeling}: $\mathcal{M}=\{\mathbf{m}_i\}$ \\
	Input LiDAR scan at timestamp $t$: $\mathcal{P}=\{\mathbf{p}_s\}$ \\
	Estimated transformation at $k-1$: $\mathbf{T}_{k-1}$
	\ENSURE 
	\STATE // Set initial guess
	\STATE $\mathbf{T}^{\text{init}} = \mathbf{T}_{k-1}$
	\STATE // Filter dense LiDAR scan
	\STATE datafilter$\left(\mathcal{P}\right)$
	\STATE // Original point-to-plane ICP
	\STATE $\mathbf{T}^\text{o},\mathcal{A}^\text{o} \leftarrow \argmin_{\mathbf{T}\in\text{SE(3)}} \left( e\left(\mathcal{M}, \mathbf{T}^{\text{init}}\mathcal{P}\right) \right)$
	\STATE // Semantic labeling by Criteria~(\ref{criteria2}) and point selection
	\STATE $\mathcal{P}^\prime \leftarrow f_{\text{select}}\left(f_{\text{label}}\left(\mathcal{P}, \mathcal{A}^o\right)\right)$
	\STATE // Weighted point-to-plane ICP with Equation~(\ref{oc},\ref{op},\ref{o})
	\STATE $\mathbf{T}_k\leftarrow \argmin_{\mathbf{T}\in\text{SE(3)}} \left( w\left(\mathcal{M}, \mathbf{T}^\text{o}\mathcal{P}^{\prime}\right) e\left(\mathcal{M}, \mathbf{T}^\text{o}\mathcal{P}^{\prime}\right) \right)$
\end{algorithmic}
\end{algorithm}

\begin{figure*}[!htb]
\centering
\subfigure[Original point-to-plane ICP]{
	\includegraphics[width=5.5cm]{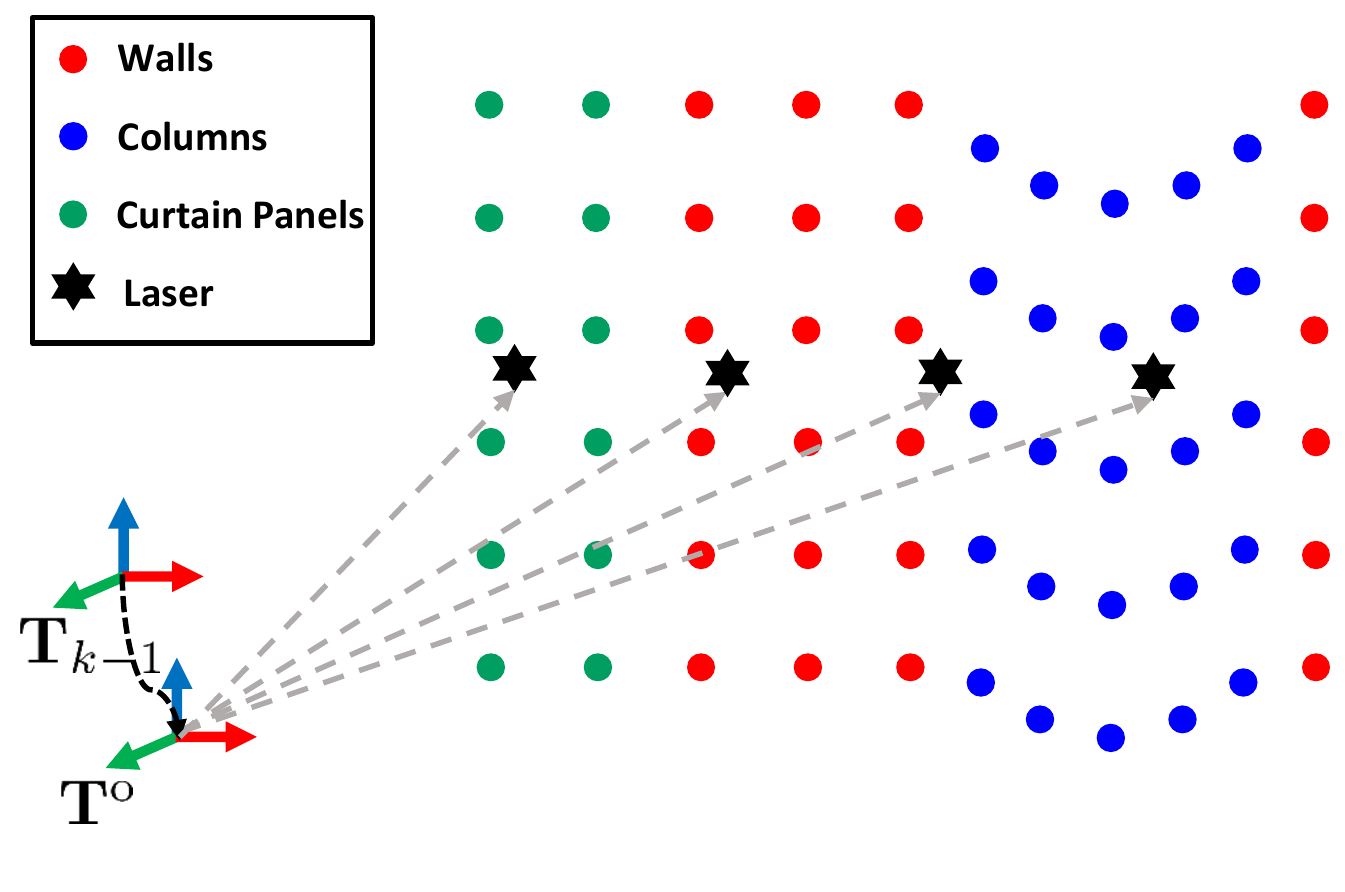}}
\subfigure[Semantic labeling and selection]{
	\includegraphics[width=5.5cm]{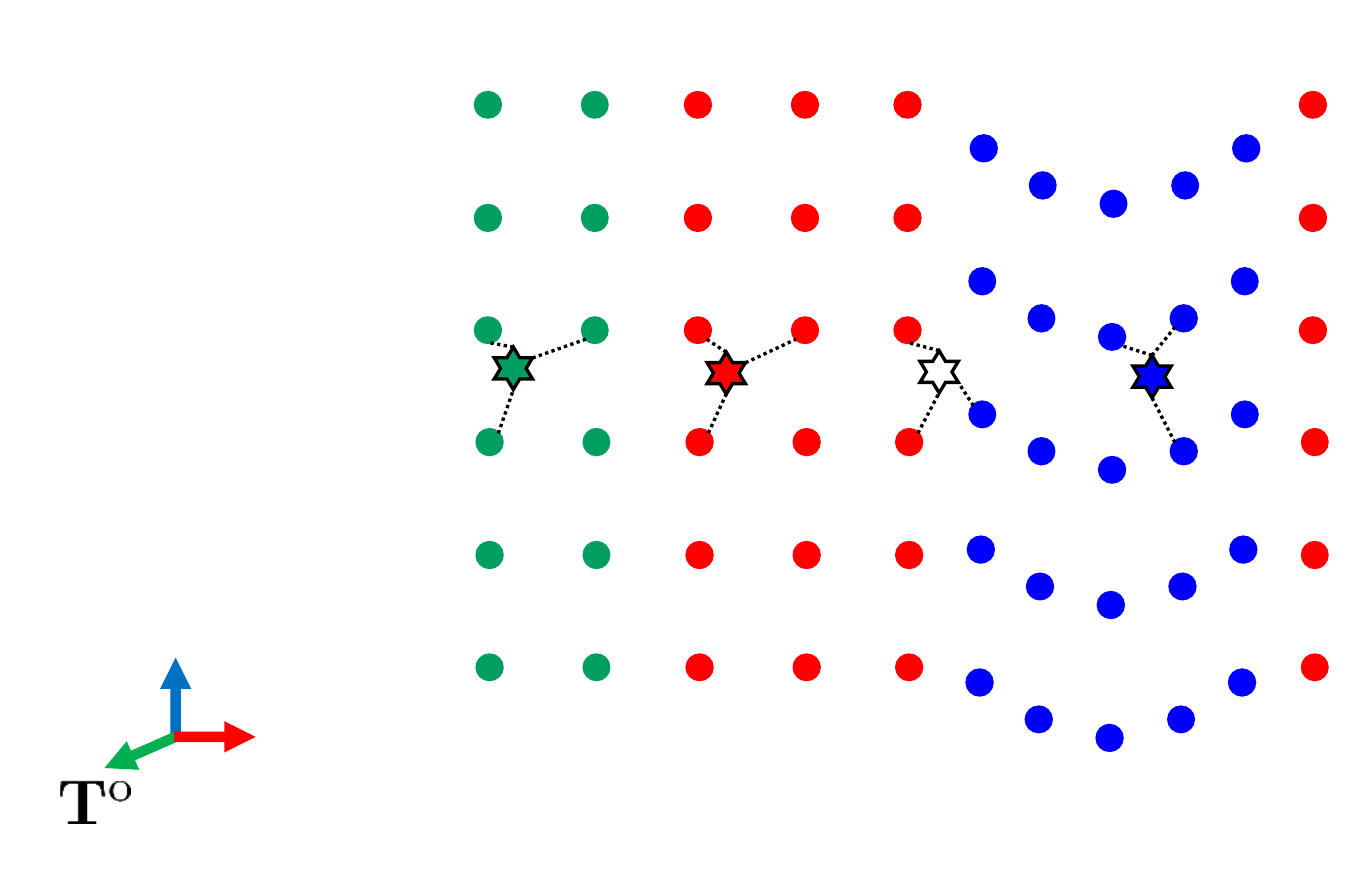}} 
\subfigure[Semantic-aided ICP]{
	\includegraphics[width=5.5cm]{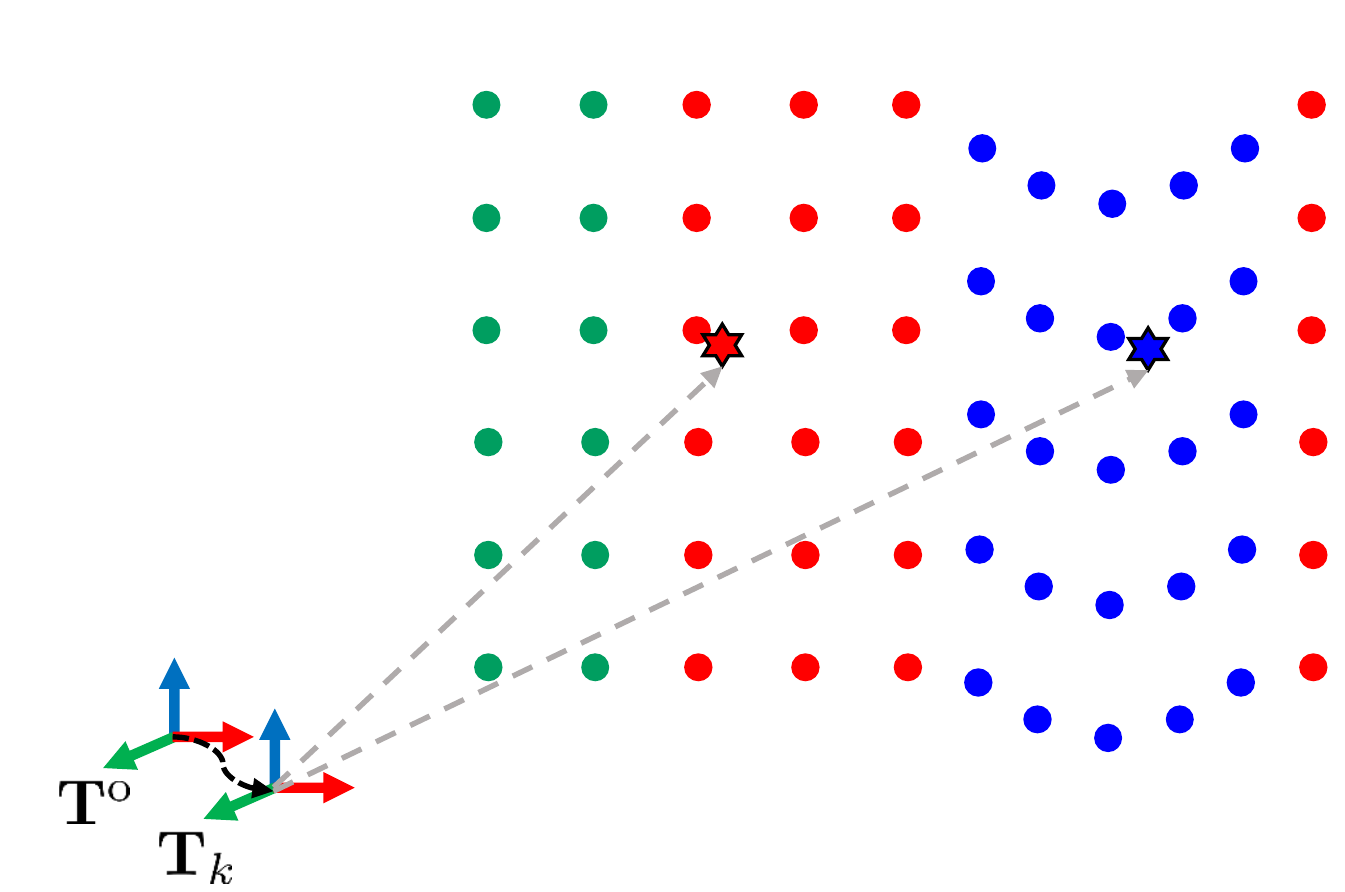}} 
\caption{A toy case for our proposed coarse-to-fine localization pipeline. (a) A point-to-plane ICP is performed as the coarse step to achieve pose estimation. (b) Then three laser points are labeled with the consistency criteria (green, red and blue), and only two are used for the fine step after selection (red and blue). (c) A semantic-aided ICP is performed as the fine step to refine robot pose at timestamp $k$.}
\label{semanticfilter}
\end{figure*}

To address this challenge, a coarse-to-fine localization is proposed and it consists of three steps: original ICP, semantic filtering,  semantic ICP. Original ICP can be regarded as the coarse step to achieve a preliminary result. Then semantic filtering step can achieve laser points labeling and selection based on the result of the first step. Finally, semantic ICP is used to refine the pose estimation. Semantic filtering and semantic ICP can be regarded as the fine step in the pipeline. The whole pipeline is illustrated in Figure~\ref{semanticloc} and Algorithm~\ref{semanticicp}.

Firstly, an original ICP is performed to minimize Equation~(\ref{error2}). The data association of the last iteration can be recorded, denoted as $\mathcal{A}^o$, as follows:

\begin{equation} \label{A}
\mathcal{A}^\text{o} =\left(
\begin{matrix}
	\left(\mathbf{p}_1, \mathbf{m}_{1,1}\right)      & \left(\mathbf{p}_2, \mathbf{m}_{2,1}\right)       & \cdots & \left(\mathbf{p}_S, \mathbf{m}_{S,1}\right)        \\
	\vdots & \vdots & \ddots & \vdots \\
	\left(\mathbf{p}_1, \mathbf{m}_{1,K}\right)      & \left(\mathbf{p}_2, \mathbf{m}_{2,K}\right)       & \cdots & \left(\mathbf{p}_S, \mathbf{m}_{S,K}\right)      \\
\end{matrix}
\right)
\end{equation}
where $S$ is the number of points in $\mathcal{P}$ and $K$ is the number of nearest neighbor search of each point in $\mathcal{P}$. Each column of $\mathcal{A}^o$ represents the matched results of one point $\mathbf{p}_s$ to its nearest neighbors after ICP alignment. 

In the second step, we check each column in $\mathcal{A}^o$ and label $\mathbf{p}_s$ if the matched map points satisfy the consistency criteria: \textit{all the matched map points should be in the same category}, formulated as follows:

\begin{equation}\label{criteria2}
c\left(\mathbf{p}_s\right) = 
\left\{ 
\begin{array}{cc}
	c\left(\mathbf{m}_{s,1}\right) \ ,& \text{if}\; c\left(\mathbf{m}_{s,1}\right) = \cdots = c\left(\mathbf{m}_{s,K}\right)
	\\
	\text{Not labeled}\ , & \text{otherwise}
\end{array}
\right.
\end{equation}

Thereafter, some points in $\mathcal{P}$ are ``labeled'' and some are not. Only labeled points are considered in the following process. However, not all the labeled points are informative, e.g., points matched as ``Windows'' may not return any meaningful measurements using LiDAR sensors. Besides that, we consider the category selection is flexible, and can be decided by users in different working environments and conditions, e.g., ``Furniture'' could be helpful for the localization when a robot is traveling in a room filled with static furniture, but might be harmful when there are many semi-dynamic office chairs.

Thus, we only select those labeled points of certain specific types, so the second step can be formulated as two steps: first label and then select, as follows:

\begin{equation} \label{secondstep}
\mathcal{P}^\prime \leftarrow f_{\text{select}}\left(f_{\text{label}}\left(\mathcal{P}, \mathcal{A}^o\right)\right)
\end{equation}
where $f_{\text{label}}\left(\cdot\right)$ and $f_{\text{select}}\left(\cdot\right)$ represent the labeling and selection process, respectively; $\mathcal{P}^\prime$ is the filtered point cloud that will be used in the following estimation.

In the third step, semantic ICP is designed to minimize Equation~(\ref{error3}) based on the coarse result from the first step. A semantic-aided weight function $w_c\left(\cdot\right)$ is formulated that incorporates the labels of LiDAR readings $\mathcal{P}^\prime$ and the map $\mathcal{M}$:

\begin{equation}\label{oc}
w_c = \left\{\begin{array}{cc}
	\mu\ , & c\left(\mathbf{p}_s\right) = c\left(\mathbf{m}_{s,k}\right) \\
	1-\mu\ , & c\left(\mathbf{p}_s\right) \neq c\left(\mathbf{m}_{s,k}\right) 
\end{array}
\right.
\end{equation}

in which $\mu\in\left[0.5, 1\right]$ is a variable that determines the importance of semantic association. If $\mu=1$, only few matched lasers are kept in challenging scenes. In the experimental section of this paper, we set $\mu=0.8$ as a constant value, which means a data association is with higher weight when the laser point is in the same category as the matched map point.

Besides the semantic-aided weight function, a Huber function \cite{huber1992robust,fitzgibbon2003robust} is also utilized to weight the data association, as follows:

\begin{equation}\label{op}
w_\rho = \left\{\begin{array}{cc}
	1\ ,  & e\left(\cdot\right) < \delta \\
	\frac{\delta}{e\left(\cdot\right)} \ ,& \text{else}
\end{array}
\right.
\end{equation}
where $\delta$ is a point-to-plane distance threshold. Finally, the overall weight function of for each matched $\left(\mathbf{p}, \mathbf{m}\right)$ is computed as the combination of the semantic and geometric relation:

\begin{equation} \label{o}
w = w_cw_\rho
\end{equation}

The proposed three-step semantic localization is shown in Figure~\ref{semanticloc} and Algorithm~\ref{semanticicp}. A toy example is also presented to illustrate the proposed coarse-to-fine localization pipeline in Figure~\ref{semanticfilter}, in which the number of nearest neighbor $K$ is set as 3 and ``Curtain Panels'' is not selected in the fine step.

To guarantee the efficiency for real-time application, we randomly sample the raw LiDAR scan and sub-sample input points in high-density regions. The initial guess of the transformation is also critical to build an efficient and robust scan matching. However, in this study, there is no Inertial Measurement Unit (IMU) or other odometry to estimate the transformation between $k-1$ and $k$, which is different from other CAD or BIM-based localization methods \cite{gao2022fp,follini2020bim,hendrikx2021connecting}. At each timestamp $k$, we set the previous estimated $\mathbf{T}_{k-1}$ as the initial guess to estimate $\mathbf{T}_k$, as shown in Figure~\ref{semanticloc} and~\ref{semanticfilter}. This means we only test the robustness and accuracy via scan-by-scan matching at the front-end, and there is no customized back-end estimator in our proposed localization pipeline.	

\begin{figure}[t]
\centering
\includegraphics[width=7cm]{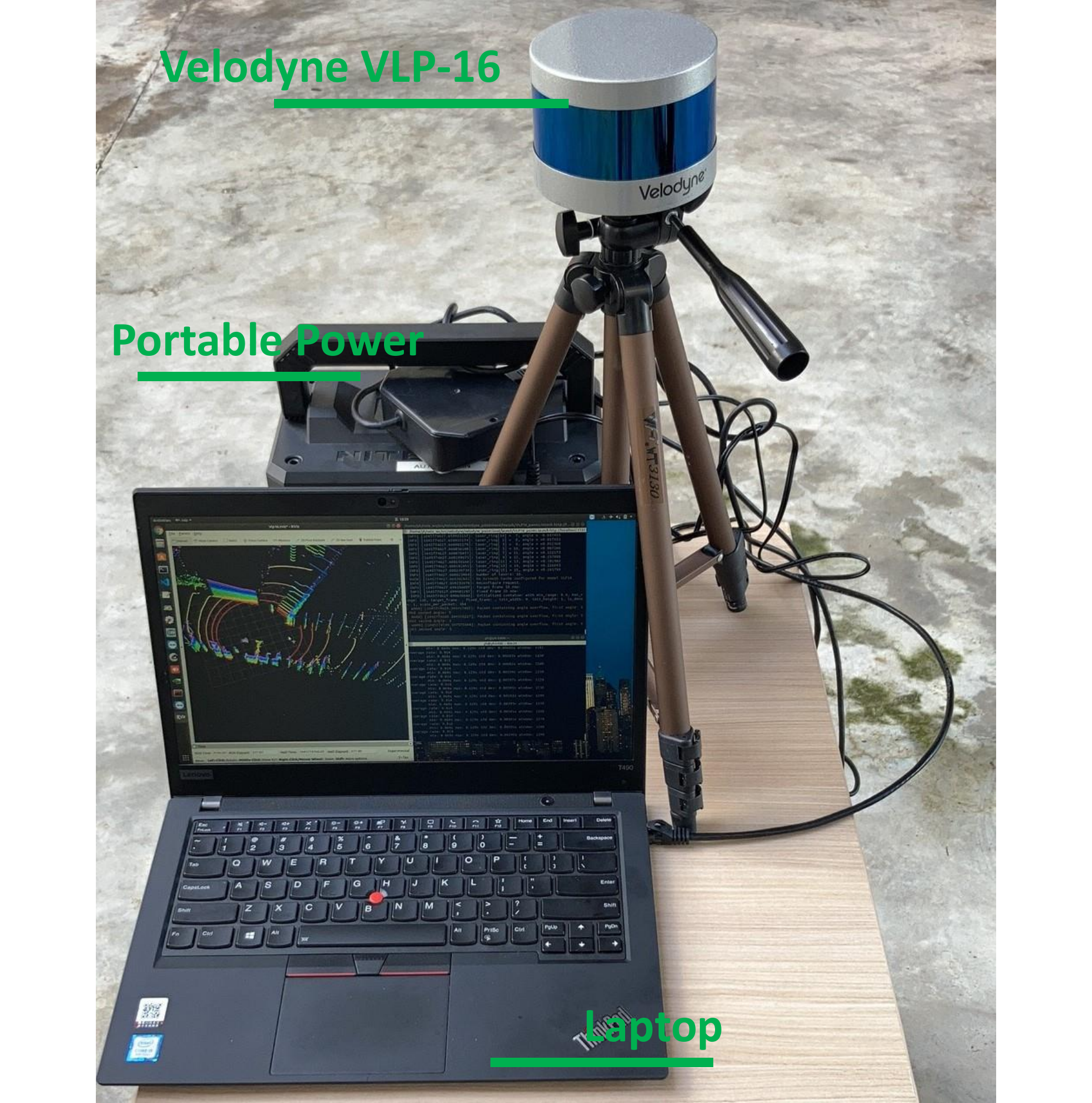}
\caption{Devices for data collection and experimental validation.}
\label{device}
\end{figure}

\section{Experiments}
\label{exp}

In order to validate the effectiveness of the proposed framework, several experiments are conducted in the real world, including the offline BIM-to-Map conversion and online semantic localization. 

\begin{table*}[t]
\captionsetup{justification=centering}
\renewcommand\arraystretch{1.5}
\begin{center}
	\caption{Self-collected Data Sequences in NUS SDE4 Building}
	\label{dataset}
	\begin{tabular}{p{2.0cm}<{\centering}p{1.0cm}<{\centering}p{3.0cm}<{\centering}p{3cm}<{\centering}p{3cm}<{\centering}}
		\hline
		\hline
		Sequence & Storey & Travel Distance (m) & \multicolumn{2}{c}{Environment}  \\
		\hline
		2-1 & 2nd & 43.5 & \multirow{3}*{\includegraphics[height=1.8cm]{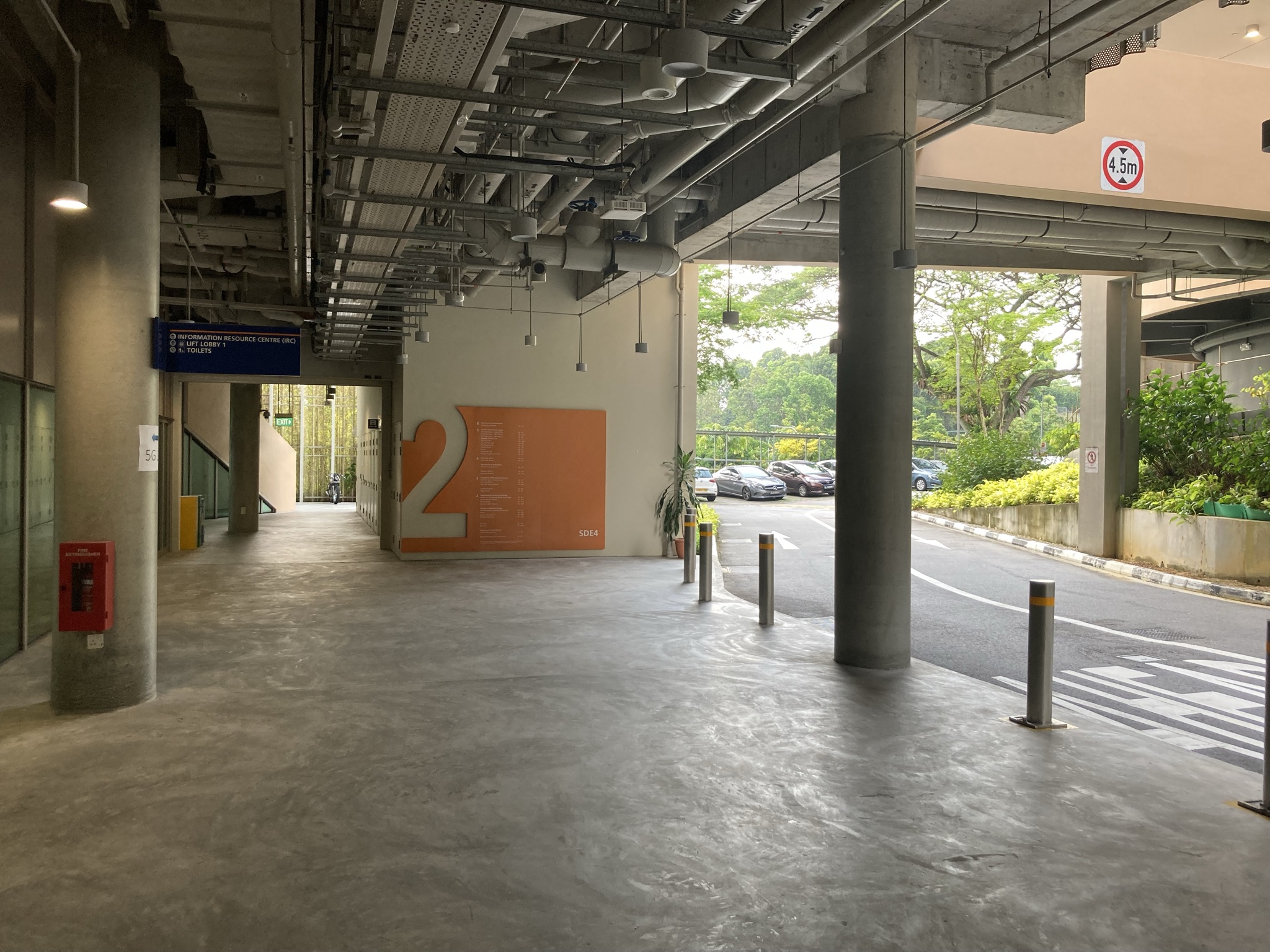}} & \multirow{3}*{\includegraphics[height=1.8cm]{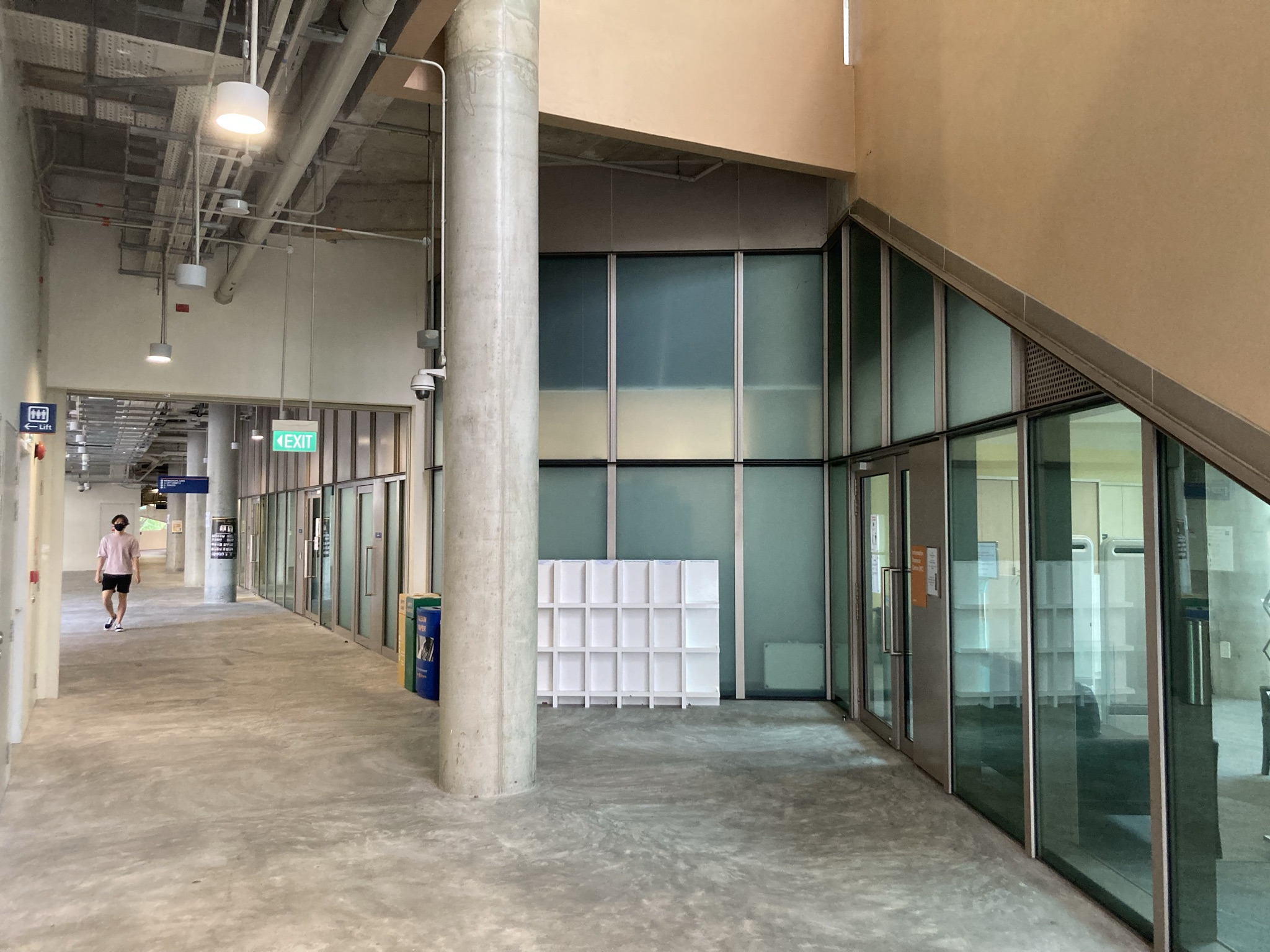}}  \\ 
		2-2 & 2nd & 39.8 & & \\
		2-3 & 2nd & 38.6 & & \\
		\hline
		3-1 & 3rd & 25.2 & \multirow{3}*{\includegraphics[height=1.8cm]{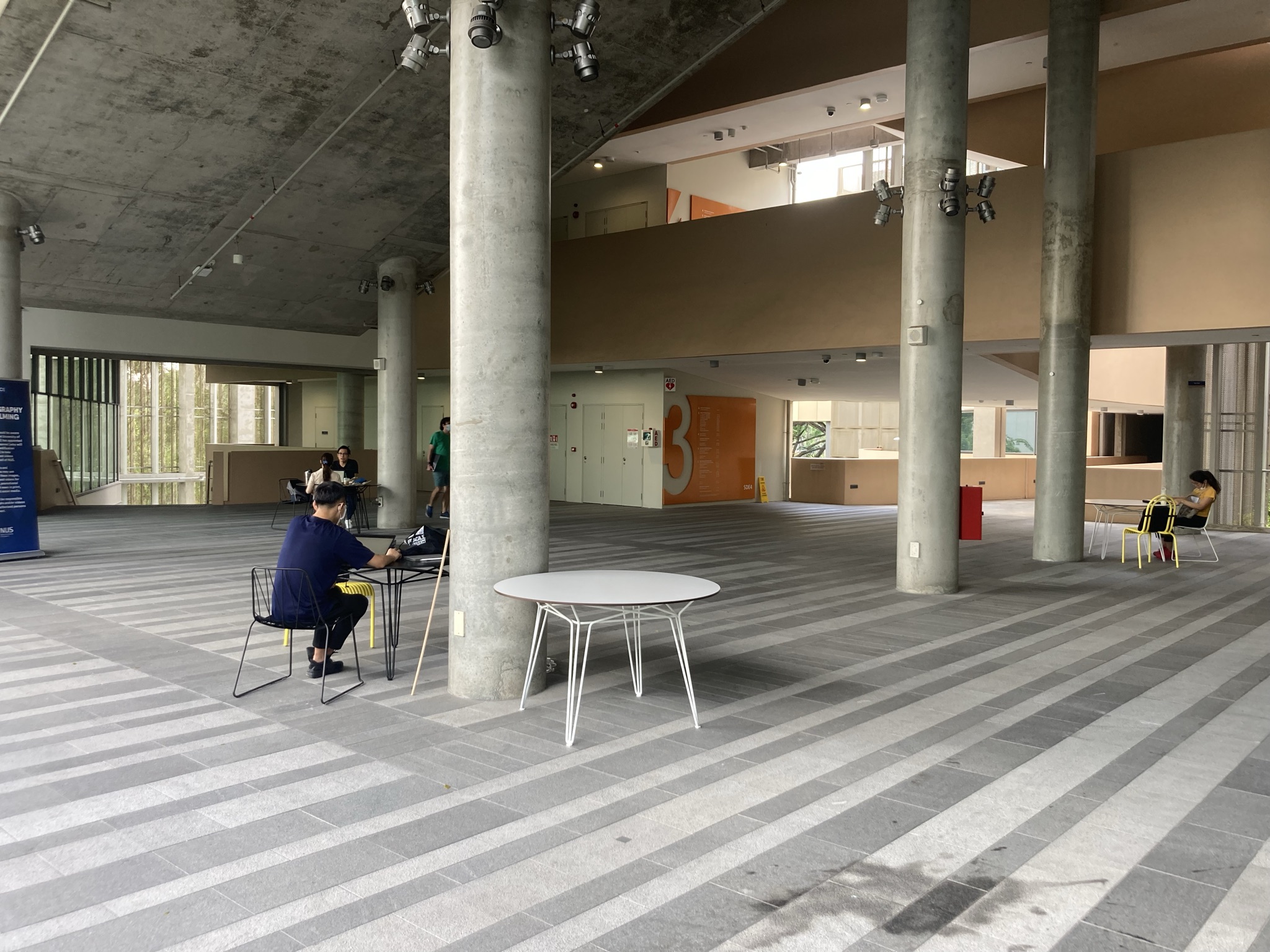}} & \multirow{3}*{\includegraphics[height=1.8cm]{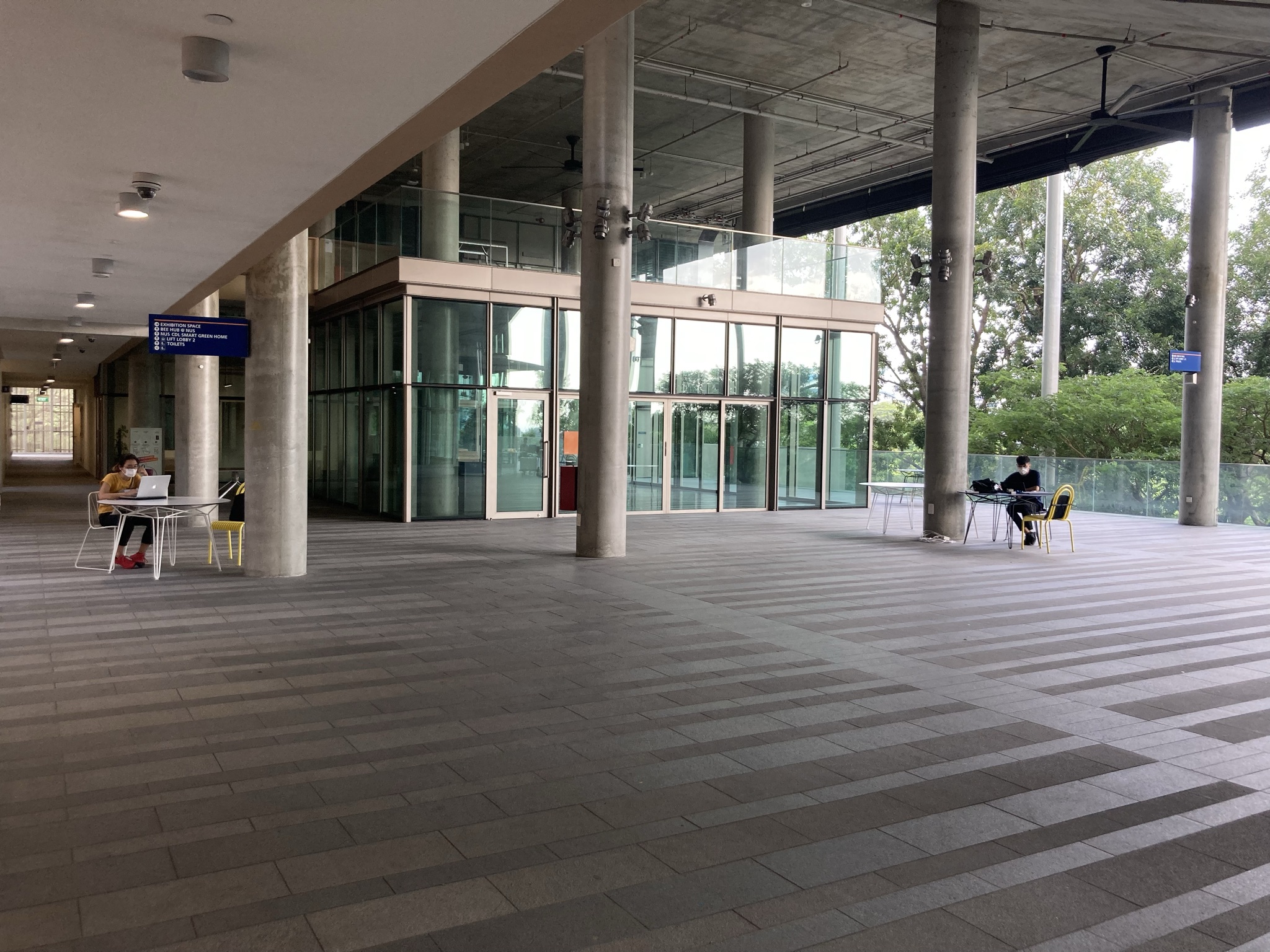}} \\ 
		3-2 & 3rd & 26.5 & & \\
		3-3 & 3rd & 48.0 & & \\
		\hline
		\specialrule{0em}{2pt}{2pt}	
		4-1 & 4th & 34.1 & \multirow{2}*{\includegraphics[height=1.8cm]{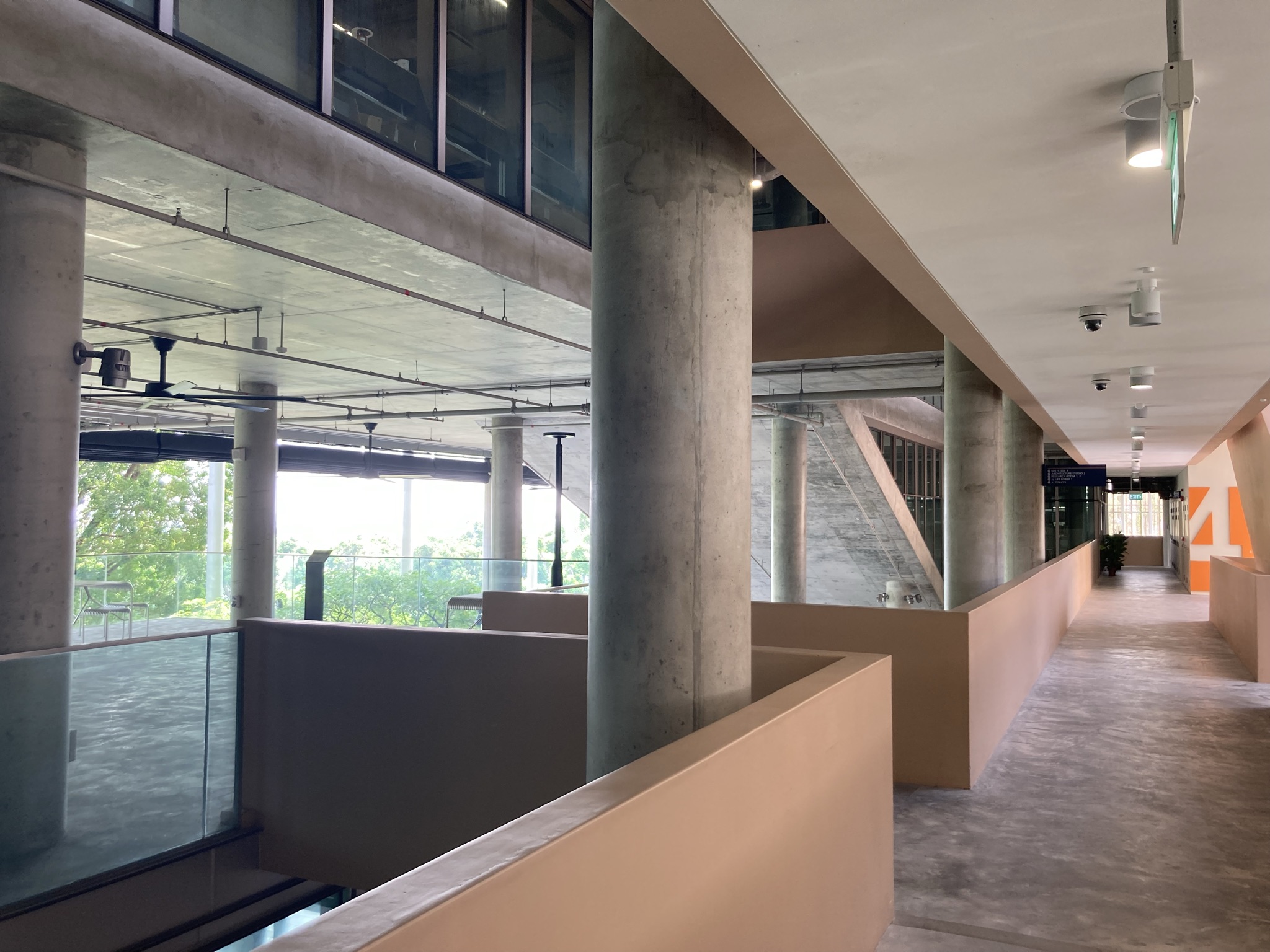}} & \multirow{2}*{\includegraphics[height=1.8cm]{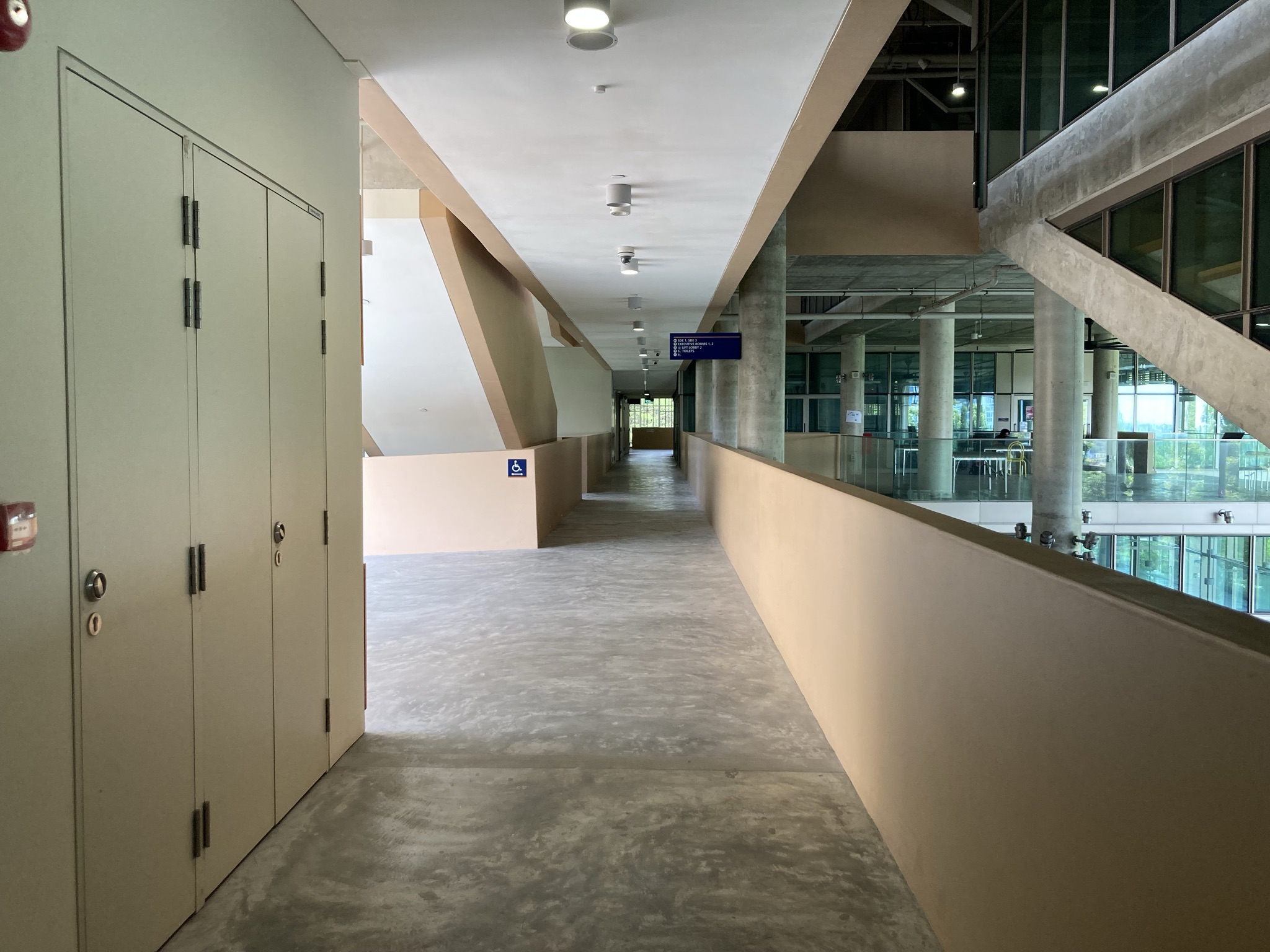}}\\ 
		\specialrule{0em}{6pt}{6pt}
		4-2 & 4th & 46.6 & & \\
		\specialrule{0em}{2pt}{2pt}
		\hline
		\specialrule{0em}{2pt}{2pt}
		5-1 & 5th & 21.1 & \multirow{2}*{\includegraphics[height=1.8cm]{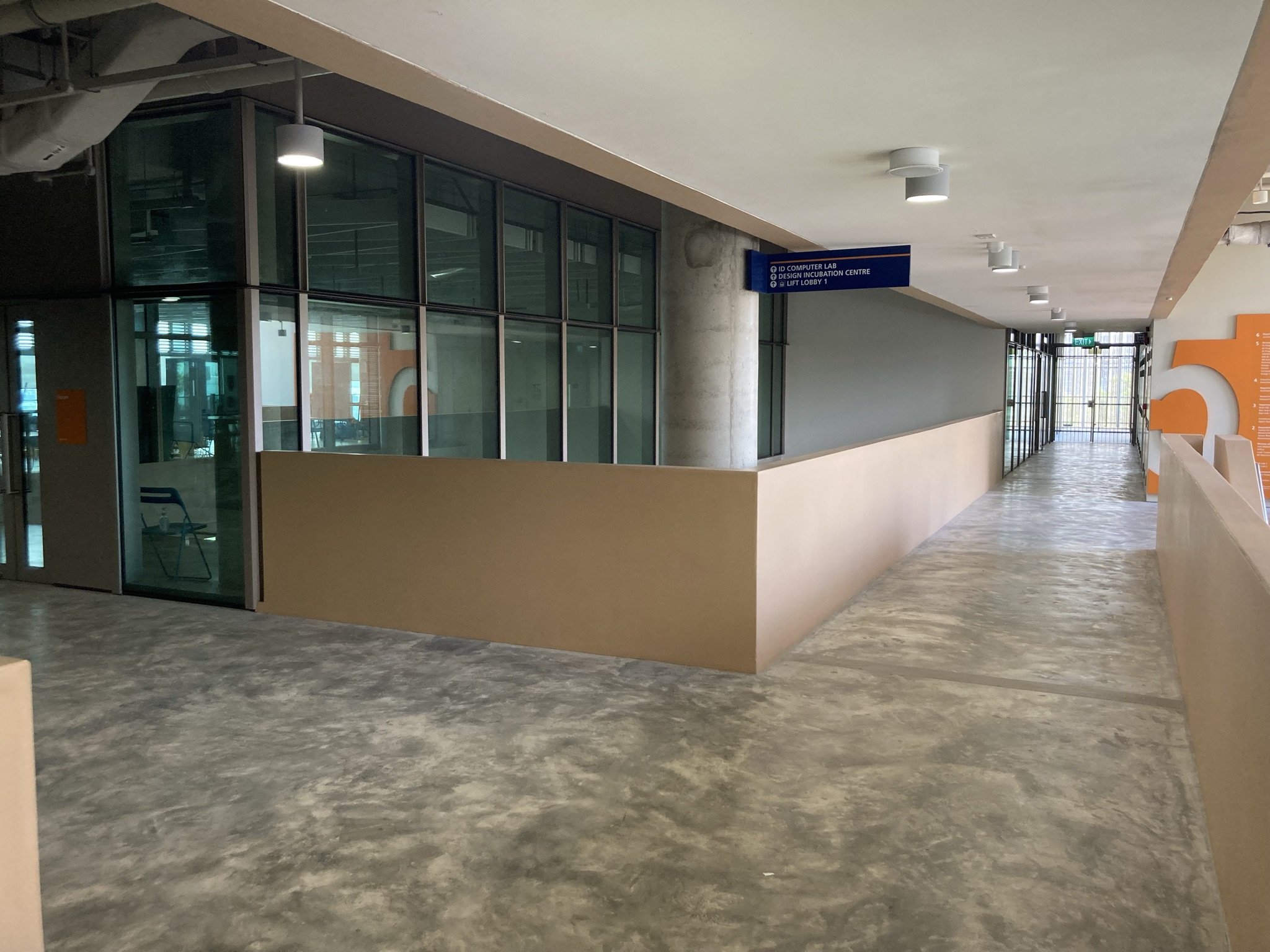}} & \multirow{2}*{\includegraphics[height=1.8cm]{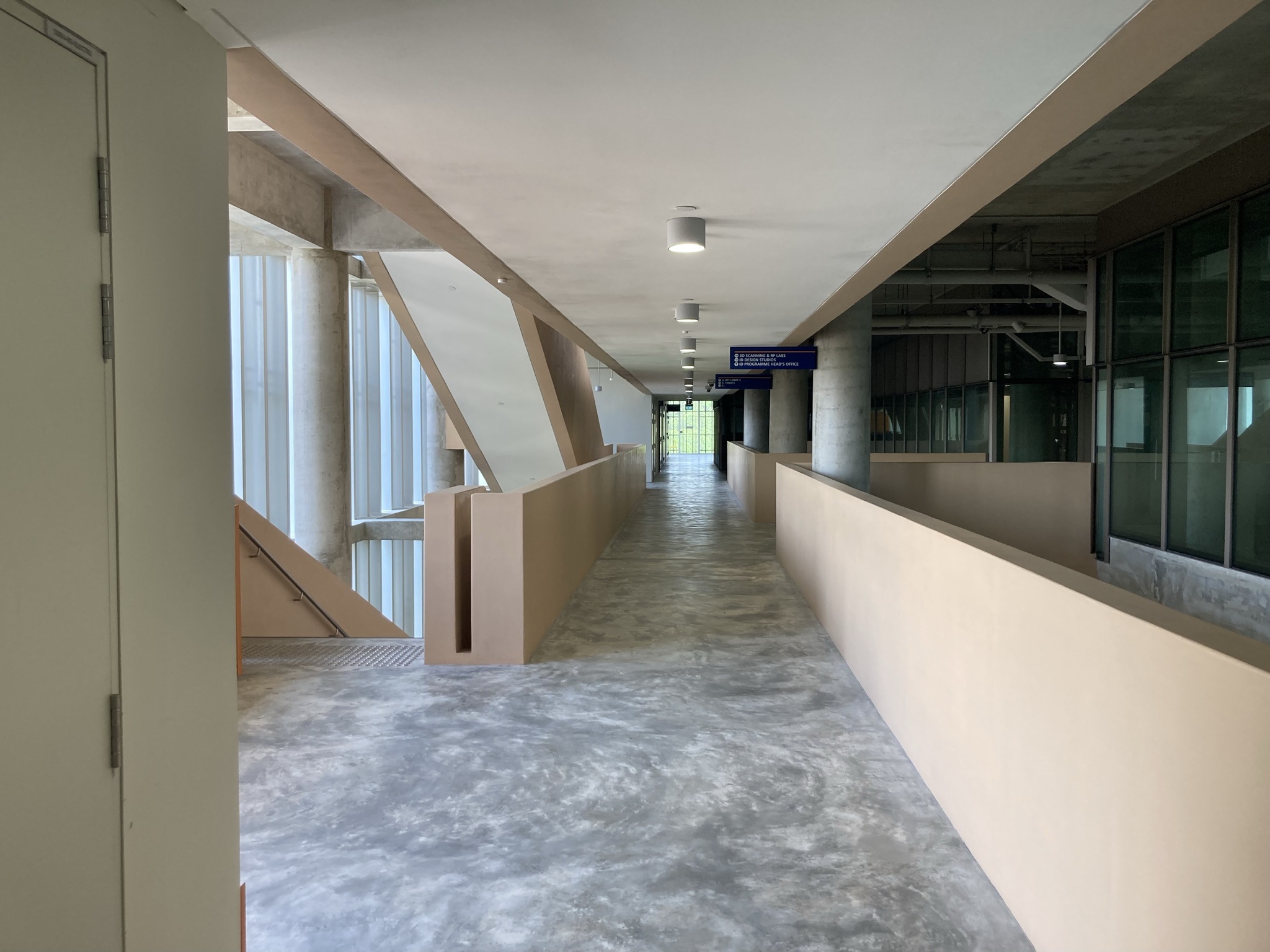}} \\ 
		\specialrule{0em}{6pt}{6pt}
		5-2 & 5th & 20.0 & & \\
		\specialrule{0em}{2pt}{2pt}
		\hline
		\hline
	\end{tabular}
\end{center}
\end{table*}

\begin{figure*}[!t]
\centering
\includegraphics[width=\linewidth]{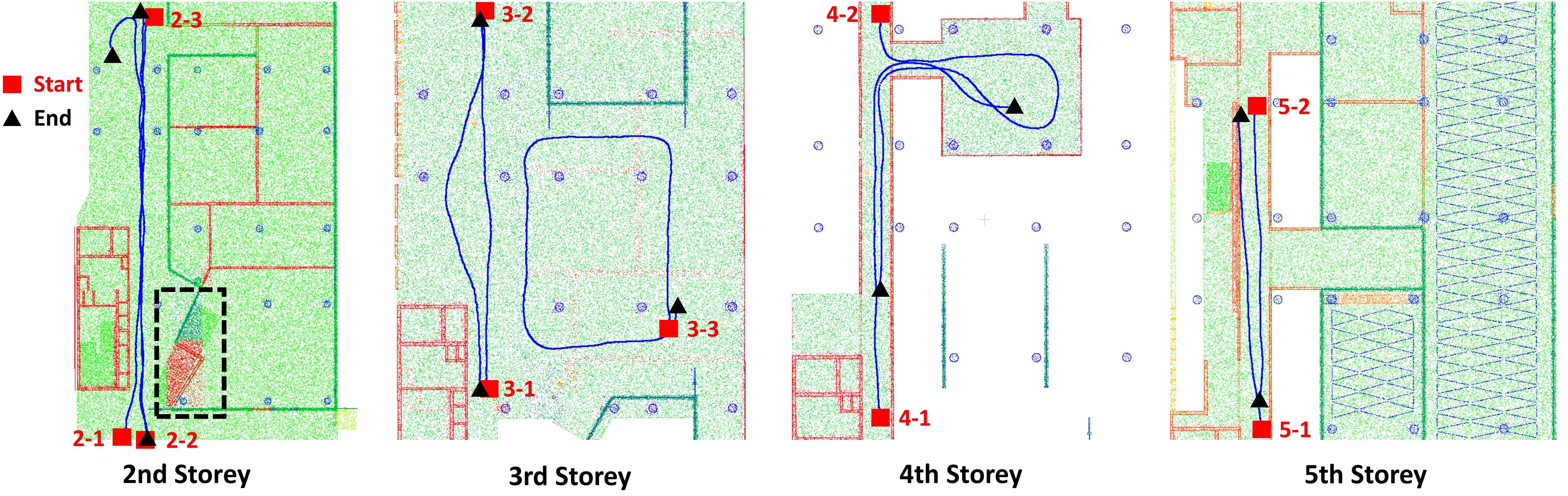}
\caption{A bird's eye view of ground truth trajectories with blue lines. Red rectangular and black triangular represent start and end position respectively. Dashed box on the 2nd Storey include incorrectly labeled map points.}
\label{sequence_traj}
\end{figure*}

\subsection{Set-up}
\label{data}

Ten sequences are collected using a Velodyne VLP-16 sensor. The data collection devices are shown in Figure~\ref{device}. All the data sessions are collected in the building of School of Design and Environment 4 (SDE4) at NUS, which is a six-storey university building. For an extensive experiment, the localization performance is tested from the second to the fifth storey, covering different environments including corridors and lounges, as shown in Table~\ref{dataset}. Approximated traveled distances of sequences are also presented. The total traveled distance is over 340 meters. 

The semantic maps are generated using Dynamo, CloudCompare and  MATLAB. The density of map points is set as 30 points/$\text{m}^3$. The semantic localization is implemented using a C++ package libpointmatcher \cite{pomerleau2013comparing} on Robot Operating System (ROS). All the online localization experiments are performed using a low power laptop with Intel I5-8265U and 16G RAM.

\begin{figure*}[t]
\centering
\includegraphics[width=\linewidth]{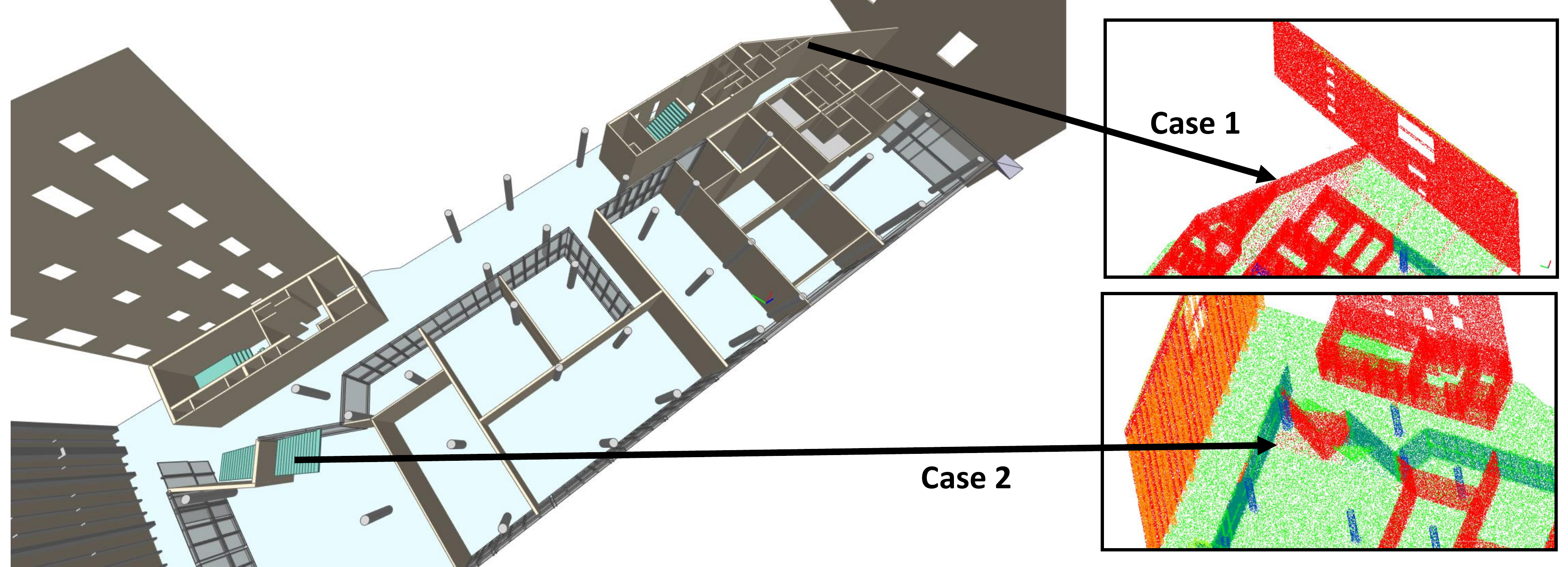}
\caption{Incorrectly or mixed labeled map points on 2nd Storey. Case 1: one wall element is aligned with a large bounding box, making some ground points are labeled with ``Wall'' (red color). Case 2: some points are aligned with two boxes.}
\label{wrong_box}
\end{figure*}

A visualization of trajectories and semantic maps are shown in Figure~\ref{sequence_traj}. Two sequences are with loop closings (Sequence 3-3 and 4-2). Furthermore, we consider that the localization tasks on the 2nd and 4th Storey are more challenging than those on the 3rd and 5th. The 2nd Storey is connected to the building entrance and the street, where a few dynamics (mostly pedestrians) exist in the collected LiDAR data. There also exist mixed and incorrectly labeled map points close to the start positions of Sequence 2-1 and 2-2, shown in Figure~\ref{sequence_traj} and~\ref{wrong_box}. The 4th Storey contains a long narrow corridor, which will degenerate the accuracy of pose tracking.

\begin{figure*}[pbt]
\centering
\subfigure[BIM and semantic map from 2nd storey to 5th storey]{ 	\label{bimandmap}
	\includegraphics[width=\linewidth]{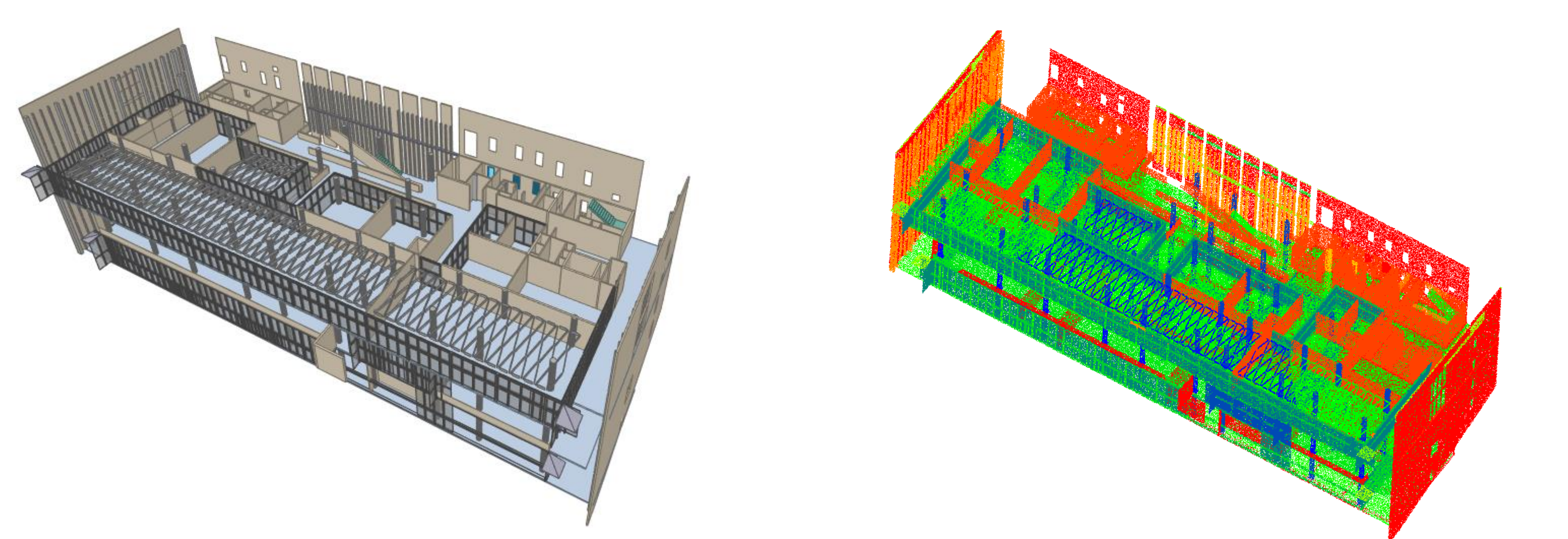}}\\
\subfigure[Map of each storey]{ \label{storeys}
	\includegraphics[width=\linewidth]{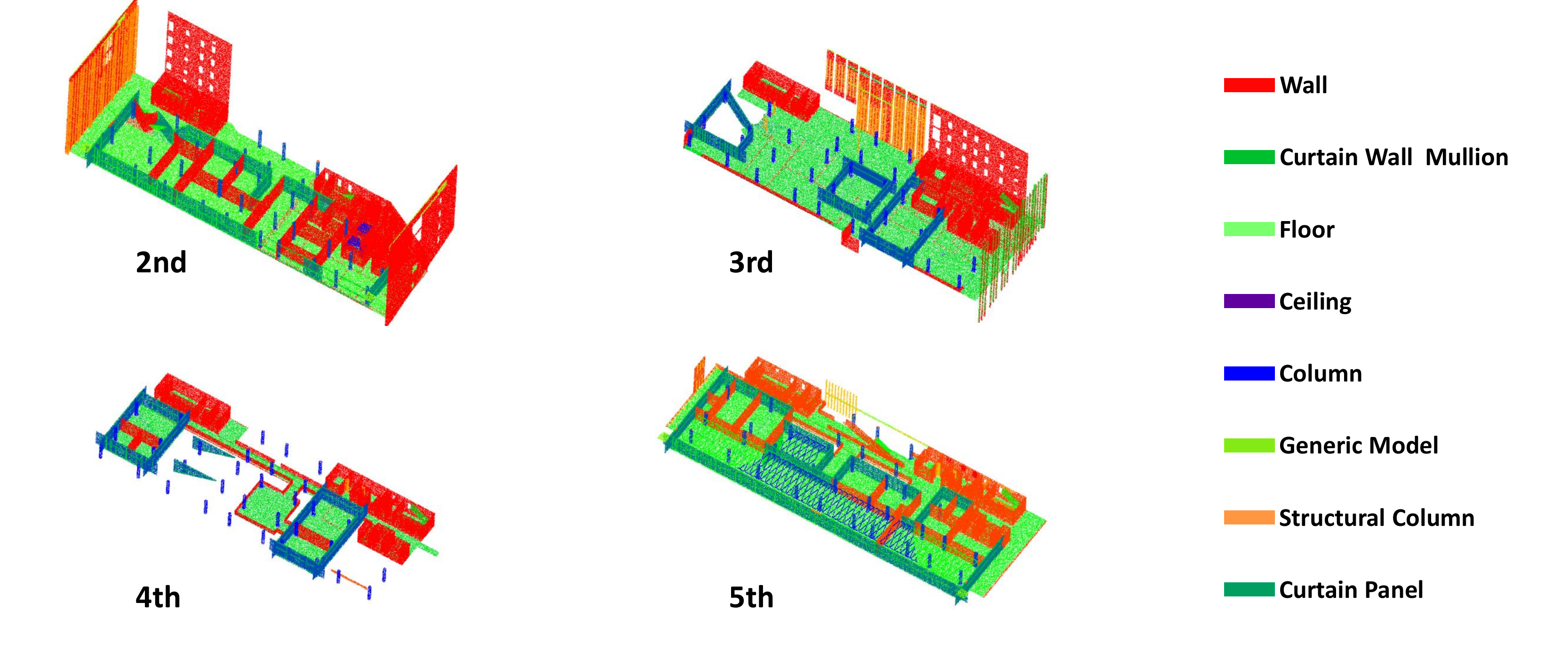}} \\
\subfigure[2nd storey]{ \label{ratio2}
	\includegraphics[width=4.25cm]{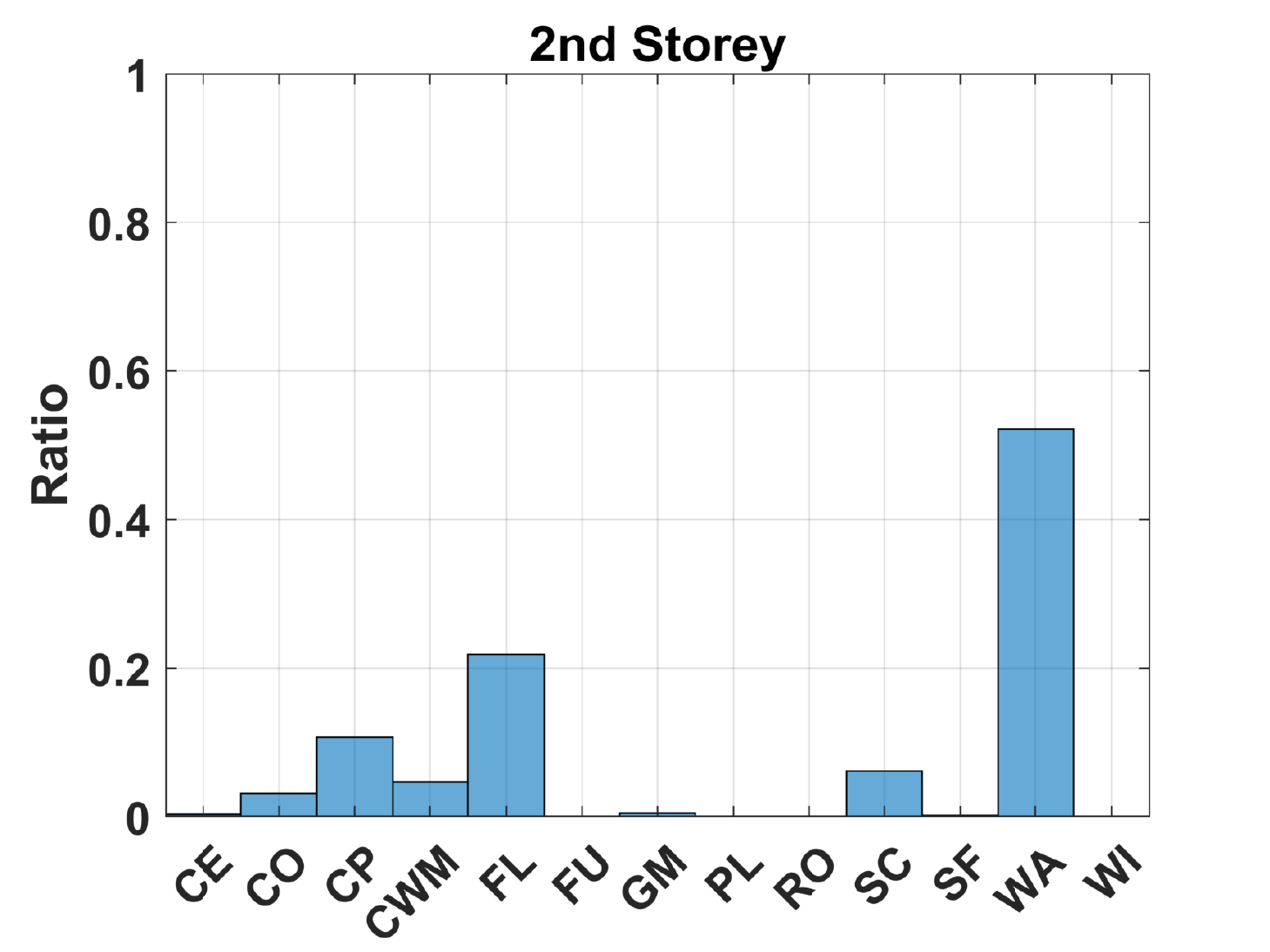}}
\subfigure[3rd storey]{ \label{ratio3}
	\includegraphics[width=4.25cm]{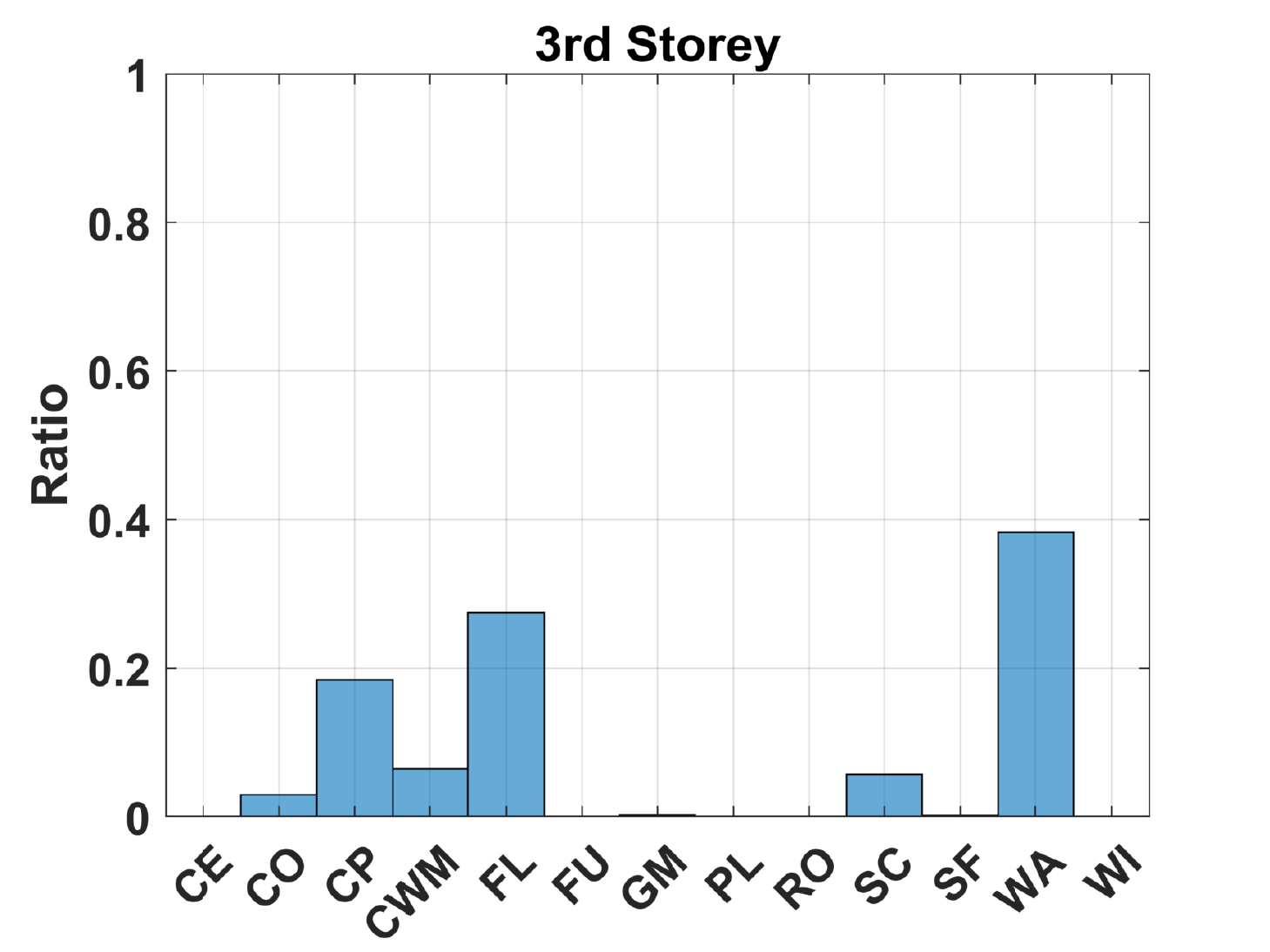}}
\subfigure[4th storey]{ \label{ratio4}
	\includegraphics[width=4.25cm]{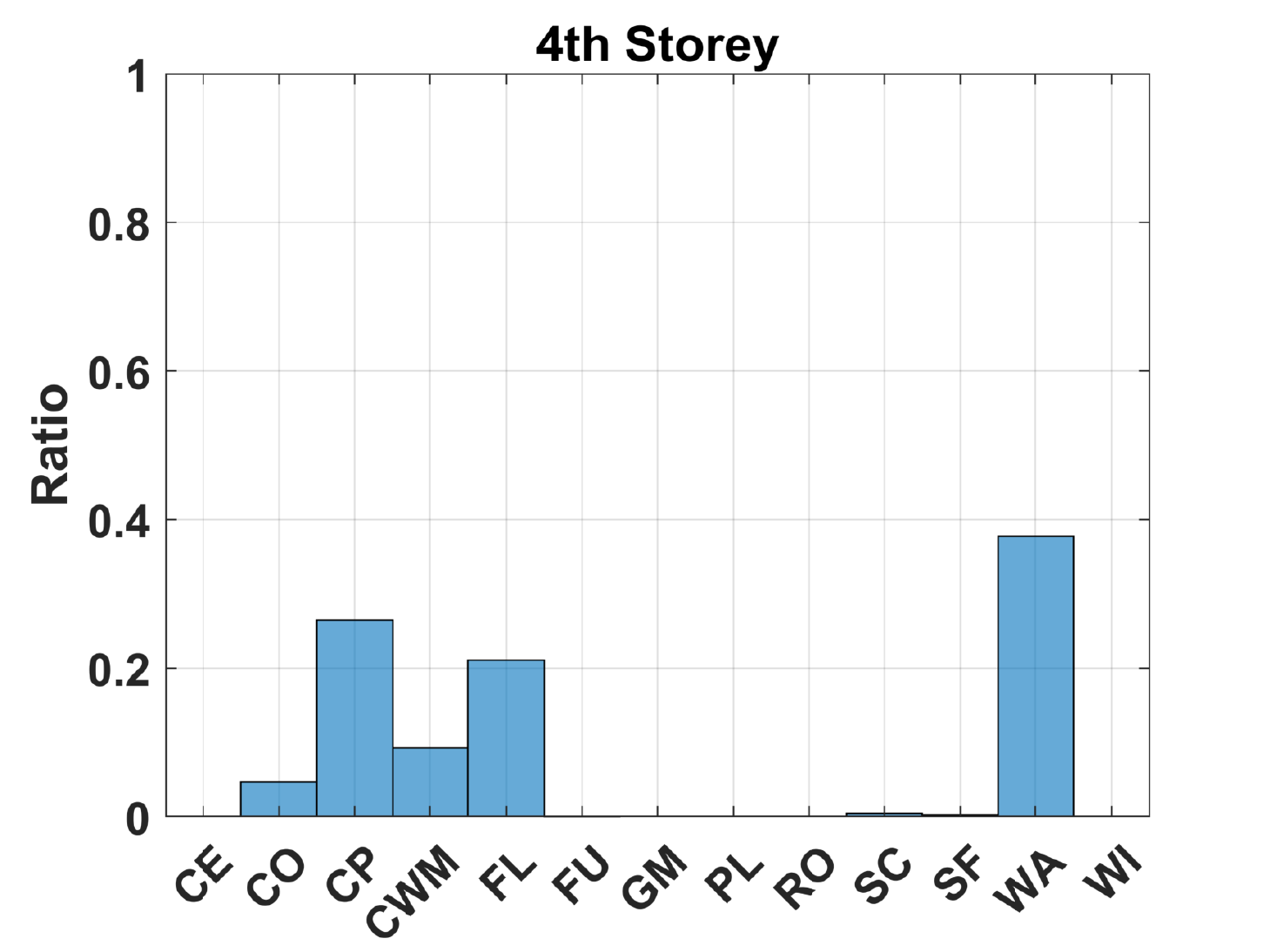}}
\subfigure[5th storey]{ \label{ratio5}
	\includegraphics[width=4.25cm]{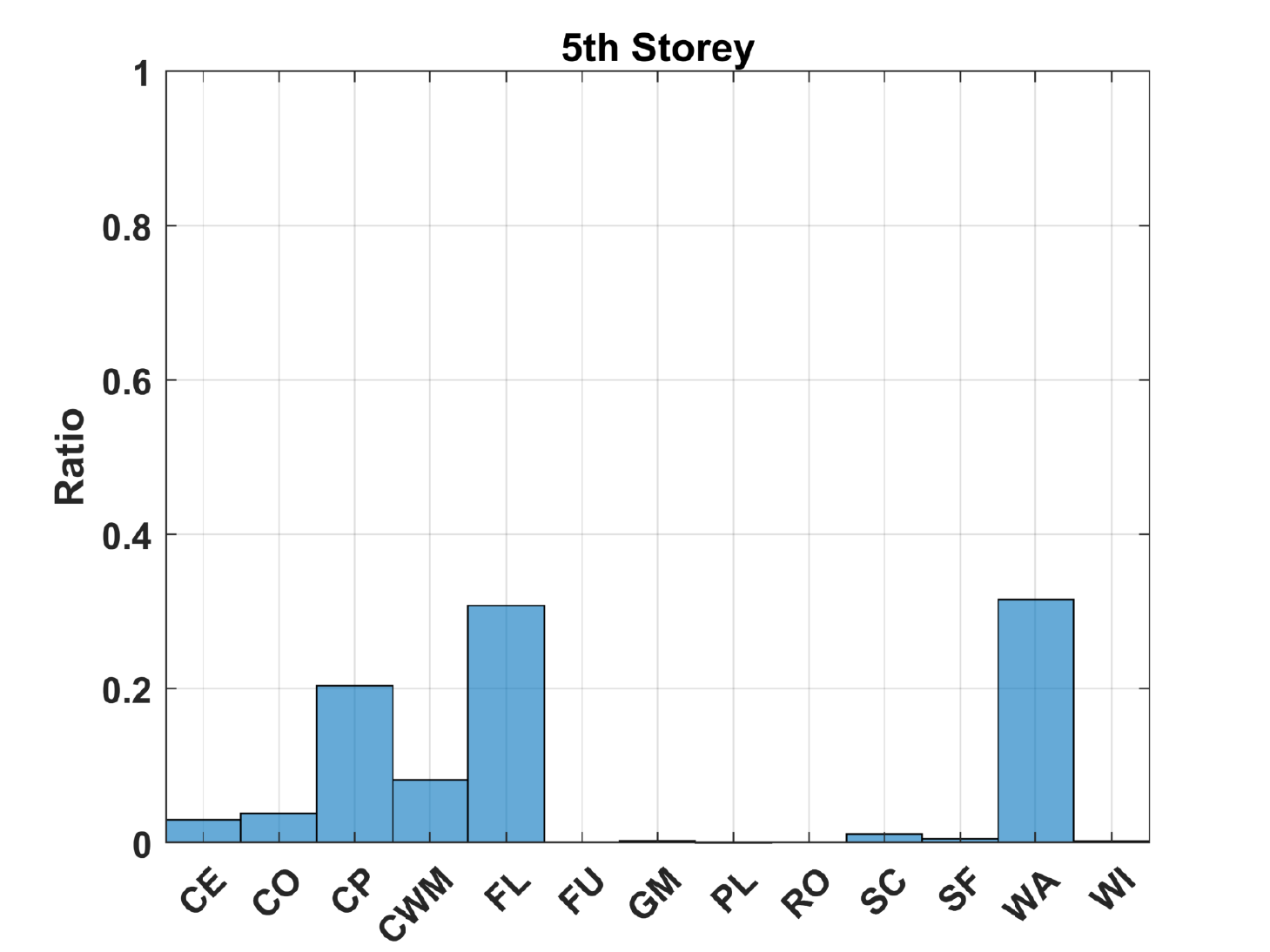}}
\caption{BIM file and semantic maps are shown in \ref{bimandmap}. Different colors indicate different categories of BIM elements in \ref{storeys}. Ratios of semantic points are also presented from \ref{ratio2} to \ref{ratio5}.}
\label{map}
\end{figure*}

\begin{figure*}[!t]
\centering
\includegraphics[width=\linewidth]{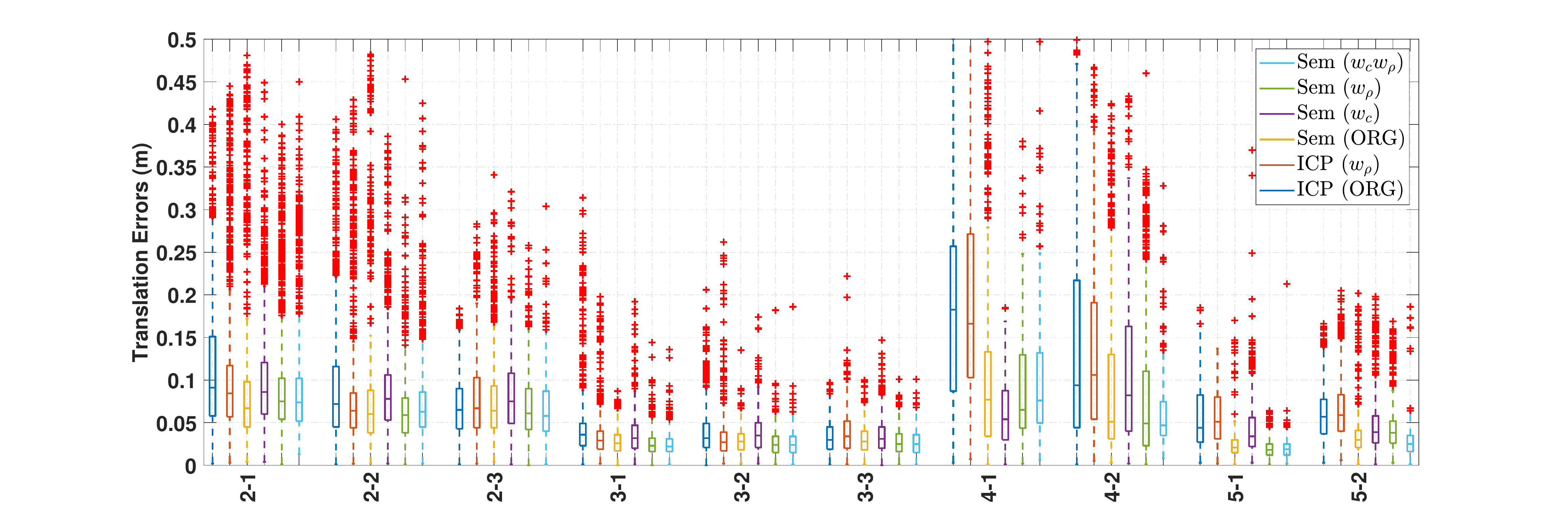}
\caption{Boxplot for visualizing summary statistics on translation errors.}
\label{error_boxplot}
\end{figure*}

\subsection{BIM-generated semantic maps}
\label{semmap}

First of all, the original BIM model and its generated maps are presented in Figure~\ref{bimandmap}. Sub-maps of individual storeys are also presented in Figure~\ref{storeys} and several categories are visualized with different colors. In SDE4 building, there are 13 categories extracted from Dynamo software: ``Ceilings (CE)'', ``Columns (CO)'', ``Curtain Panels (CP)'', ``Curtain Wall Mullions (CWM)'', ``Floors (FL)'', ``Furniture (FU)'', ``Generic Models (GM)'', ``Planting (PL)'', ``Roofs (RO)'', ``Structural Columns (SC)'', ``Structural Framing (SF)'', ``Walls (WA)'' and ``Windows (WI)''. Furthermore, we make a statistic on the number of each category and the distributions are shown in Figure~\ref{ratio2},\ref{ratio3},\ref{ratio4} and \ref{ratio5}. 

As shown in Figure~\ref{storeys}, the mapping pipeline can generate semantically augmented point cloud maps. With regards to the distribution of points, it was found that most points are associated to Walls ($\approx 40\%$), Floors ($\approx 20\%$) and Curtain Panels ($\approx 20\%$). We find that there are some incorrectly or mixed labeled map points on the 2nd storey, shown in Figure~\ref{wrong_box}. The main reasons for this problem have been analyzed in Section~\ref{bimtomap}. On the other hand, most of the map points are labeled with correct categories. In the next subsection, the proposed semantic localization pipeline will be evaluated in the NUS SDE4 building quantitatively.


\begin{table*}[t]
\captionsetup{justification=centering}
\renewcommand\arraystretch{1.5}
\begin{center}
	\caption{RMSE of Sem (ORG) tests on Sequence 3-1 with different combinations of BIM elements}
	\label{point_selection}
	\begin{tabular}{p{1.2cm}<{\centering}|p{2.0cm}<{\centering}p{2.0cm}<{\centering}p{3.0cm}<{\centering}p{1.5cm}<{\centering}p{1.5cm}<{\centering}p{2.0cm}<{\centering}}
		\hline
		\hline
		- & ALL CPNT & FL+WI+CP & FL+WI+CP+WA & FL+WA & FL+CO & FL+WA+CO \\
		\hline
		Tr. (m) & 0.062 & 0.139 & 0.085 & 0.071 & 0.037 & \textbf{0.030} \\
		Rt. ($^\circ$) & 0.540 & 0.730 & 1.048 & 0.817 & \textbf{0.337} & 0.385 \\
		\hline
		\hline
	\end{tabular}
\end{center}
\end{table*}

\begin{table*}[!t]
\captionsetup{justification=centering}
\renewcommand\arraystretch{1.5}
	\begin{center}
		\caption{RMSE of Each Sequence}
		\label{rmse}
		\begin{tabular}{p{0.8cm}<{\centering}|p{0.8cm}<{\centering}p{0.8cm}<{\centering}|p{0.8cm}<{\centering}p{0.8cm}<{\centering}|p{0.8cm}<{\centering}p{0.8cm}<{\centering}|p{0.8cm}<{\centering}p{0.8cm}<{\centering}|p{0.8cm}<{\centering}p{0.8cm}<{\centering}|p{0.8cm}<{\centering}p{0.8cm}<{\centering}}
			\hline
			\hline
			\multirow{2}*{Seq.} & \multicolumn{2}{c|}{ICP (ORG)} & \multicolumn{2}{c|}{ICP ($w_\rho$)} & \multicolumn{2}{c|}{Sem (ORG)} & \multicolumn{2}{c|}{Sem ($w_c$)} & \multicolumn{2}{c|}{Sem ($w_\rho$)} & \multicolumn{2}{c}{Sem ($w_cw_\rho$)}  \\
			&Tr.(m) &Rt.($^\circ$) &Tr.(m) &Rt.($^\circ$) &Tr.(m) &Rt.($^\circ$) &Tr.(m) &Rt.($^\circ$) &Tr.(m) &Rt.($^\circ$) &Tr.(m) &Rt.($^\circ$) \\
			\hline
			2-1 & 0.140 & 0.780 & 0.139 & \textbf{0.668} & 0.127 & 0.976 & 0.117 & 0.891 & 0.115 & 0.891 & \textbf{0.113} & 0.893 \\
			2-2 & 0.116 & 0.657 & 0.112 & \textbf{0.536} & 0.121 & 0.919 & 0.107 & 0.814 & \textbf{0.081} & 0.603 & 0.097 & 0.759 \\
			2-3 & 0.078 & 0.640 & 0.091 & 0.734 & 0.091 & \textbf{0.441} & 0.093 & 0.739 & 0.079  & 0.611 & \textbf{0.077} & 0.590 \\
			\hline
			3-1 & 0.062 & 0.540 & 0.041 & 0.409 & 0.030 & 0.385 & 0.044 & 0.425 & \textbf{0.029} & 0.348 & \textbf{0.029} & \textbf{0.335} \\
			3-2 & 0.049  & 0.485 & 0.045 & 0.387 & 0.032 & 0.375 & 0.044 & 0.417 & \textbf{0.030} & 0.343 & \textbf{0.030} & \textbf{0.324}  \\
			3-3 & 0.037 & 0.367 & 0.044 & 0.442 & 0.034 & \textbf{0.348} & 0.041 & 0.372 & 0.031 & 0.362 & \textbf{0.030} & 0.359 \\
			\hline
			4-1 & 0.235 & 1.284 & 0.292 & 1.406 & 0.179 & 0.681 & \textbf{0.076} & 1.389 & 0.127 & \textbf{0.671} & 0.137  & 0.842 \\
			4-2 & 0.193 & 1.097 & 0.236 & 2.029 & 0.129 & 0.792 & 0.136 & 0.989 & 0.117 & \textbf{0.651} & \textbf{0.079} & 1.198 \\
			\hline
			5-1 & 0.069 & 0.366 & 0.066 & 0.722 & 0.031 & 0.404 &  0.058 & 0.407 & \textbf{0.022} & \textbf{0.327} & 0.025 & 0.369 \\
			5-2 & 0.072 & 0.539 & 0.078 & 0.548 & 0.046  & 0.365 & 0.061 & 0.479 & 0.054 & \textbf{0.318} & \textbf{0.036} & 0.345  \\
			\hline
			All & 0.122 & 0.735 & 0.143 & 0.956 & 0.100 & 0.640 & 0.091 & 0.721 & 0.081 & \textbf{0.573} & \textbf{0.080} & 0.663 \\
			\hline
			\hline
		\end{tabular}
	\end{center}
\end{table*}

\begin{figure*}[t]
\centering
\subfigure[Sem ($w_cw_\rho$) localization on Sequence 4-1]{
	\includegraphics[width=9cm]{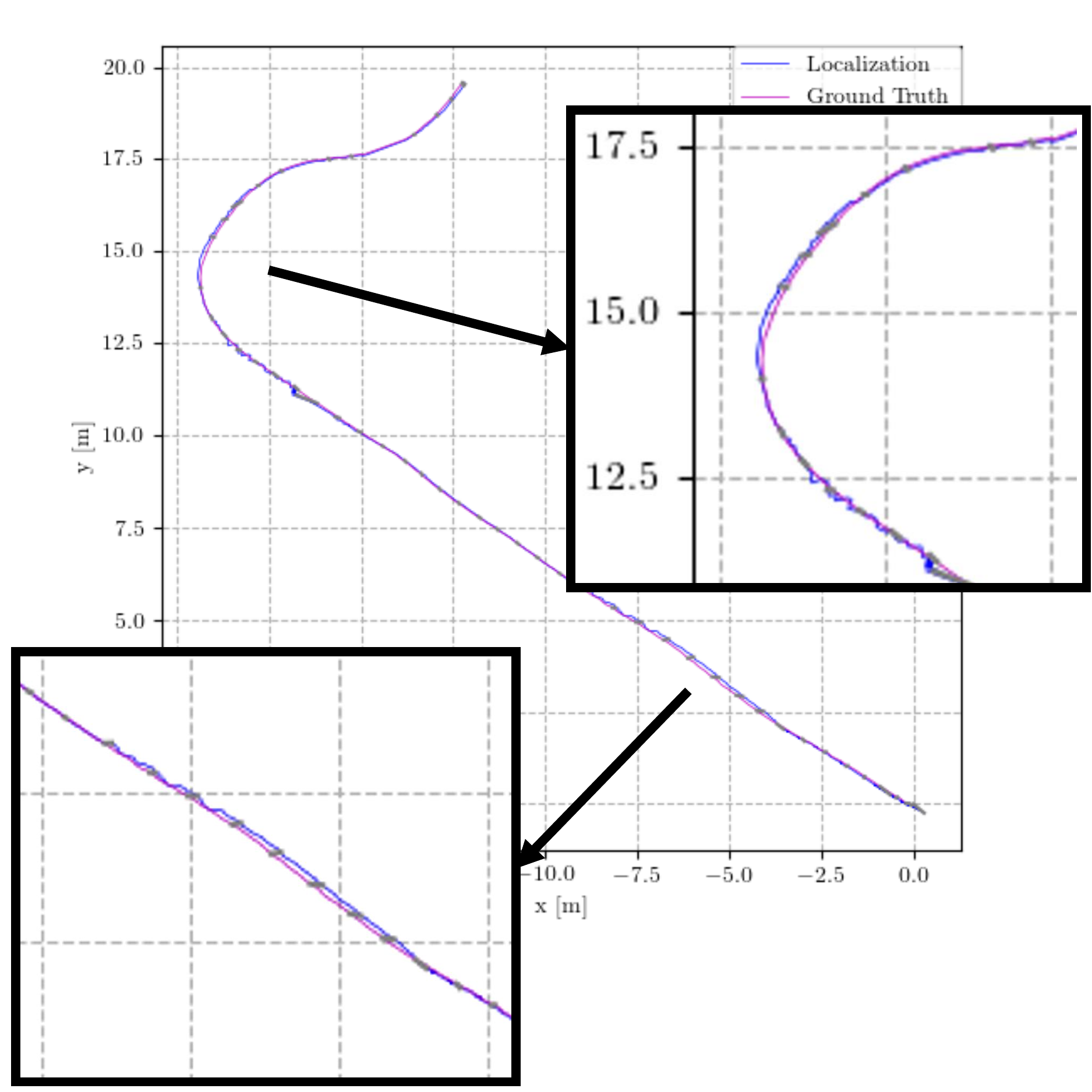}}
\subfigure[Sem ($w_cw_\rho$) localization on Sequence 5-1]{
	\includegraphics[width=9cm]{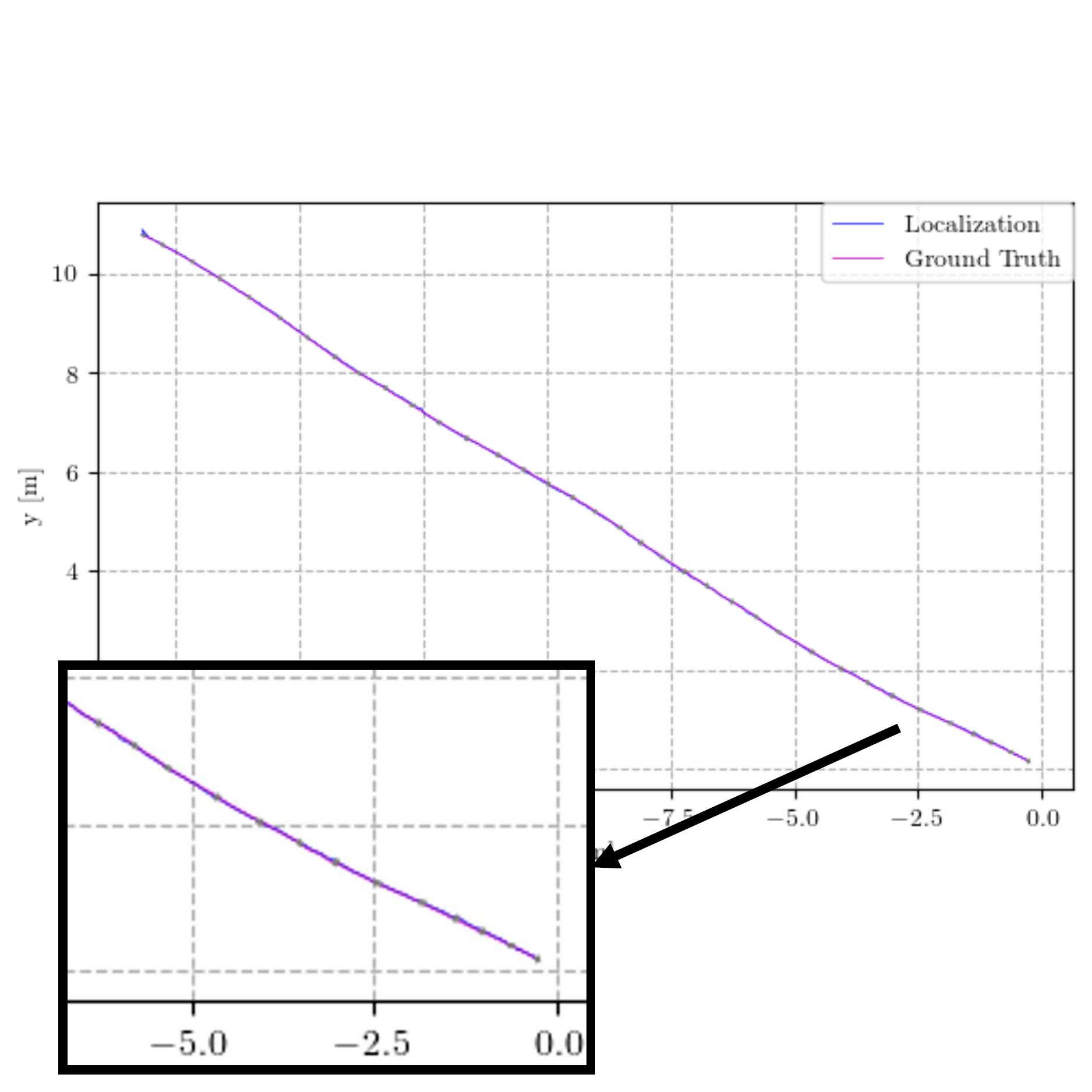}}
\caption{Trajectory comparison between ground truth and localization using \cite{zhang2018tutorial}. Gray lines are aligned connections between two poses with same sensor timestamps.}
\label{Trajectories}
\end{figure*}

\begin{figure*}[!t]
\centering
\subfigure[ICP (ORG)]{
	\includegraphics[width=5.5cm]{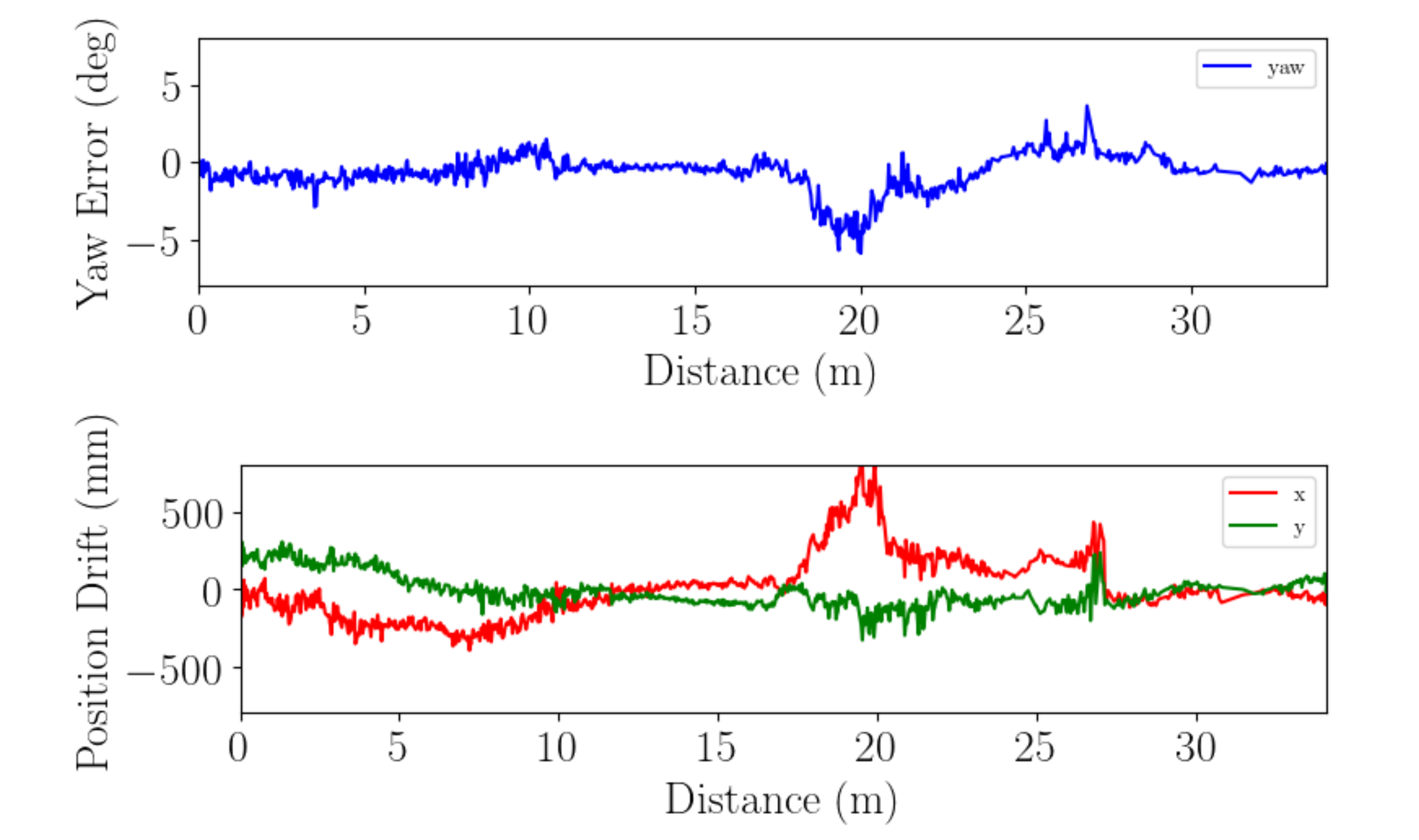}}
\subfigure[Sem (ORG)]{
	\includegraphics[width=5.5cm]{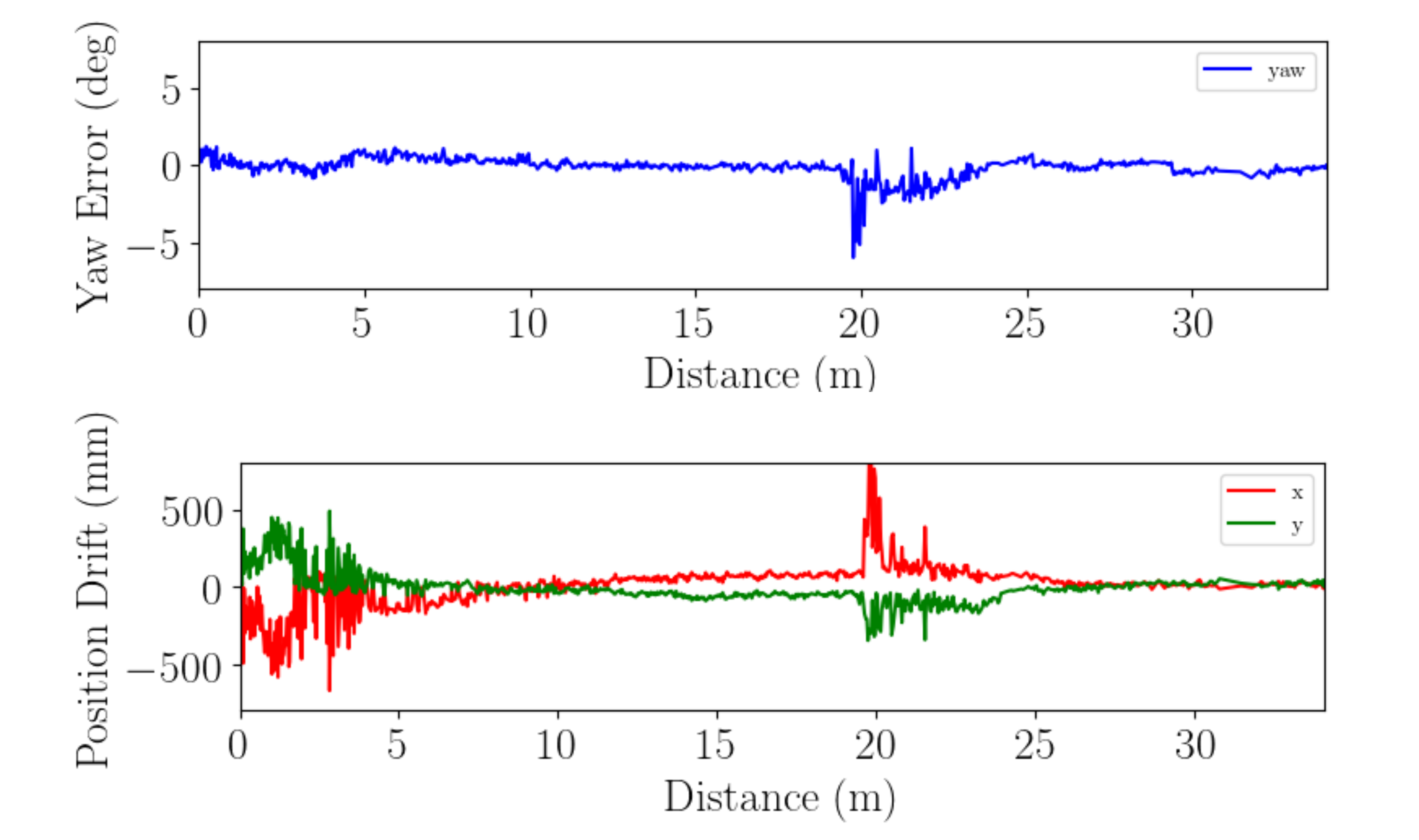}}
\subfigure[Sem ($w_cw_\rho$)]{
	\includegraphics[width=5.5cm]{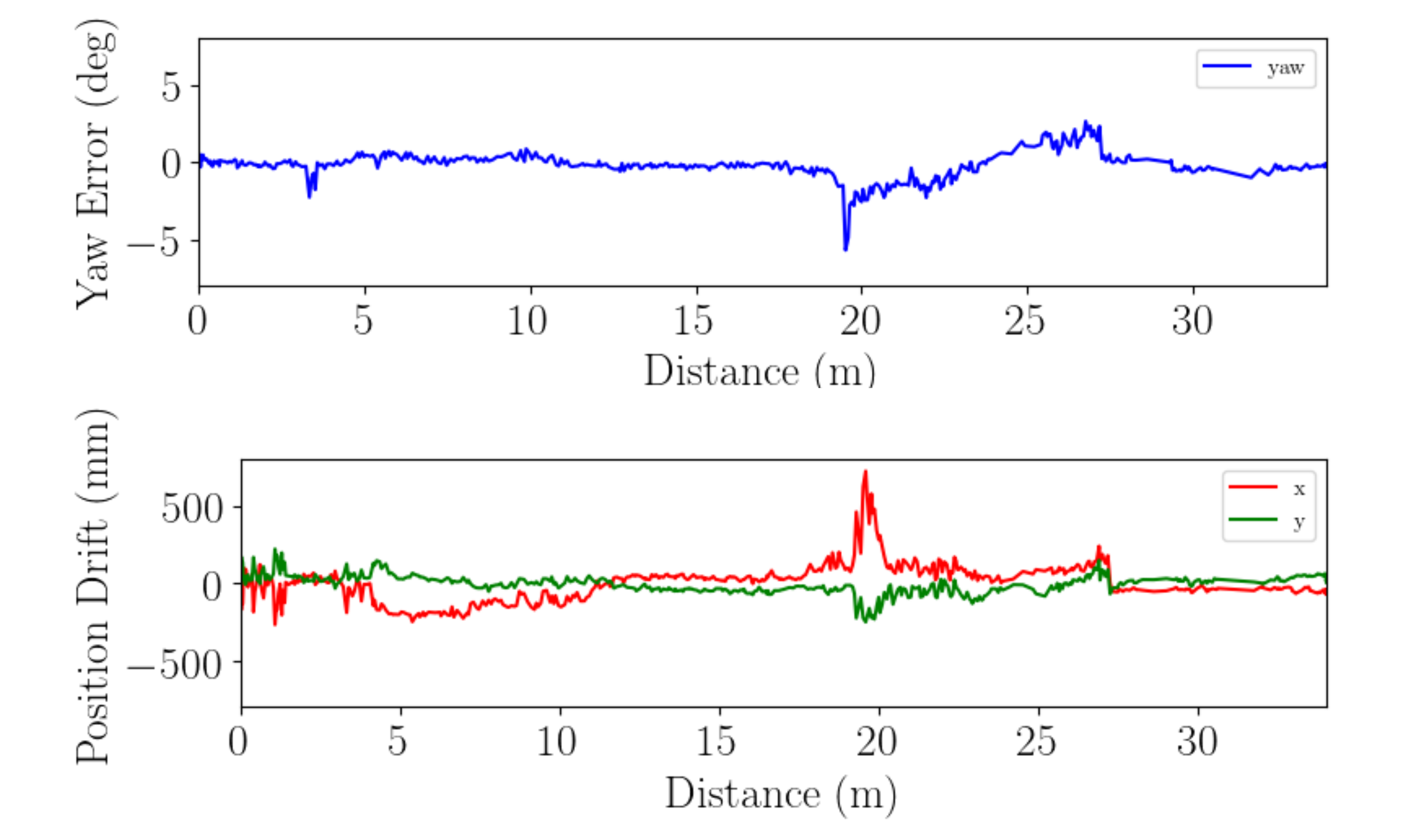}}
\caption{Error variation on Sequence 4-1. Red, green and blue lines represent errors of x, y and yaw estimation with respect to traveled distances.}
\label{traj4-1}
\end{figure*}

\subsection{Localization evaluation}
\label{loc}

Ground truth poses are required to evaluate the continuous localization. But compared to outdoor autonomous vehicles equipped with GPS/INS, it is challenging to collect ground truth poses in indoor scenes, since motion capture systems, such as Vicon, are difficult to be deployed across all halls and corridors in a large building.

In recent research work \cite{chen2020sloam,nie2021forest}, state-of-the-art SLAM methods are often used as a proxy for ground truth during evaluation. Cartographer \cite{hess2016real} is a well-designed SLAM system with a loop closing module. In \cite{zou2021comparative}, it showed a superior performance over other SLAM systems in indoor scenes, and we adopt this as ground truth. Specifically, in this paper, Cartographer is run with our fine-tuned parameters and also with low-speed \emph{rosbag} to generate ground truth poses. Cartographer and some other SLAM methods typically require IMU sensors to achieve accurate 3D pose estimation, which is infeasible with only a mobile LiDAR sensor in this study. Thus we set Cartographer with 2D configurations and evaluate our method in 2D space ($x$,$y$ and $yaw$). 

The generated ground truth poses are not aligned to map reference or BIM model. To obtain localization errors, we set sensor timestamps as indexes in all the trajectories, and utilize the open source tool \cite{zhang2018tutorial} to achieve trajectory alignment and error calculation. The Root Mean Square Error (RMSE) is calculated as follows:

\begin{equation} \label{rmse1}
\text{RMSE}_{\text{Trans.}} = \sqrt{\frac{1}{N}\sum_{n=1}^{N}(\Delta \mathbf{t}_n)^2}
\end{equation}
\begin{equation} \label{rmse2}
\text{RMSE}_{\text{Rot.}} = \sqrt{\frac{1}{N}\sum_{n=1}^{N}(\Delta \mathbf{R}_n)^2}
\end{equation}

in which $\Delta \mathbf{t}$ and $\Delta \mathbf{R}$ are the translation error ($x$ and $y$) and rotation error (only $yaw$ angle) between one estimated pose and ground truth pose respectively.

\begin{figure*}[!t]
\centering
\subfigure[ICP (ORG)]{
	\includegraphics[width=5.5cm]{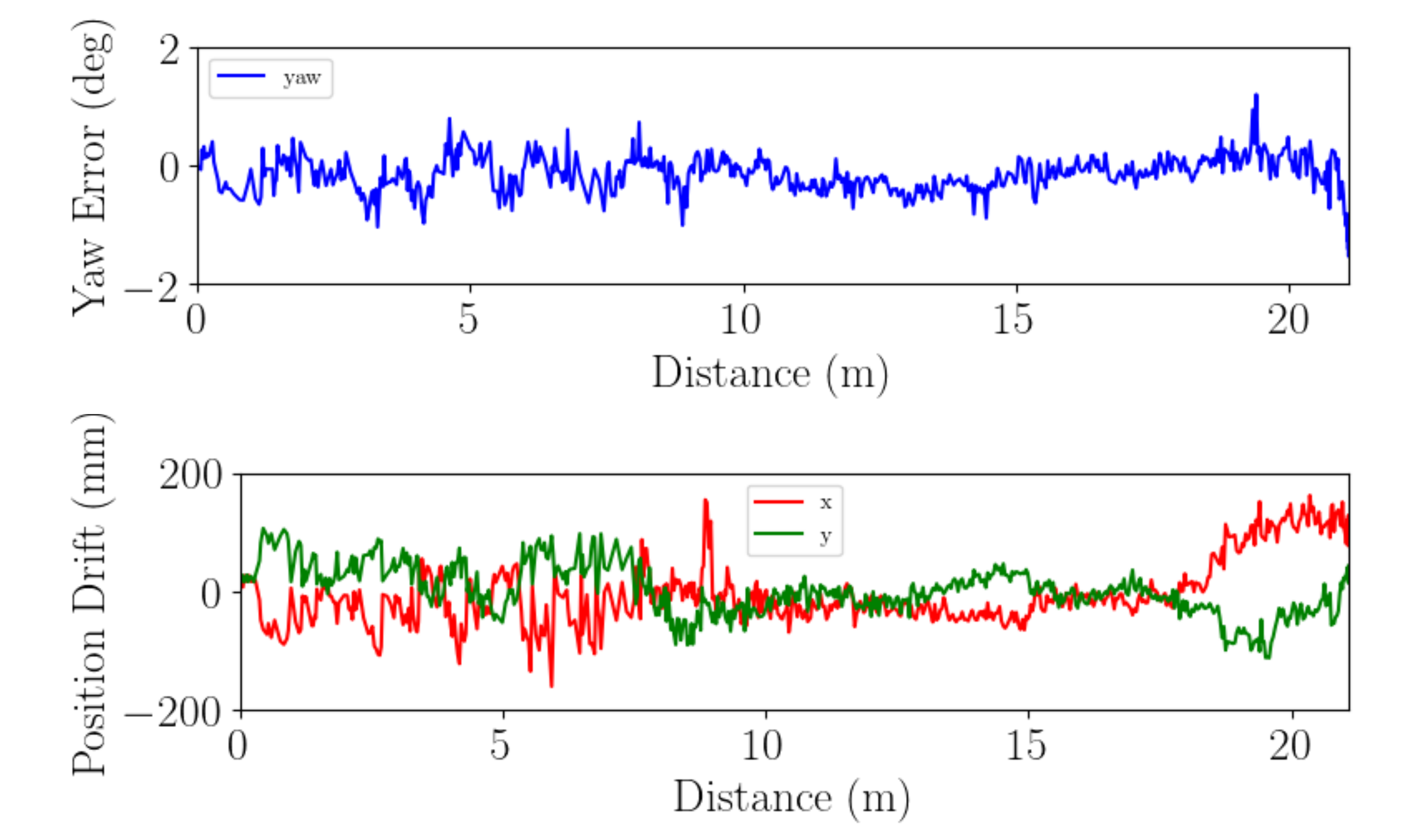}}
\subfigure[Sem (ORG)]{
	\includegraphics[width=5.5cm]{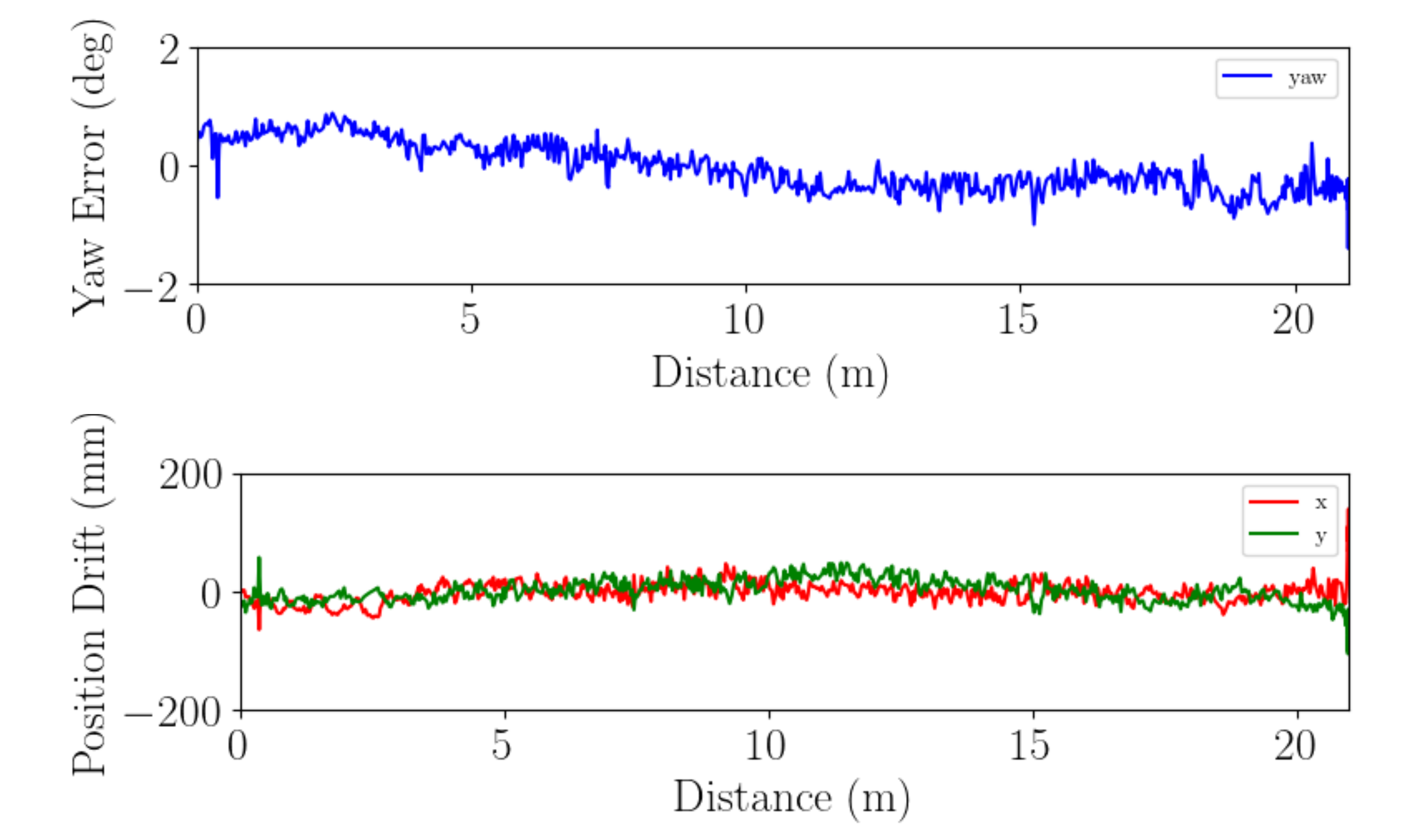}}
\subfigure[Sem ($w_cw_\rho$)]{
	\includegraphics[width=5.5cm]{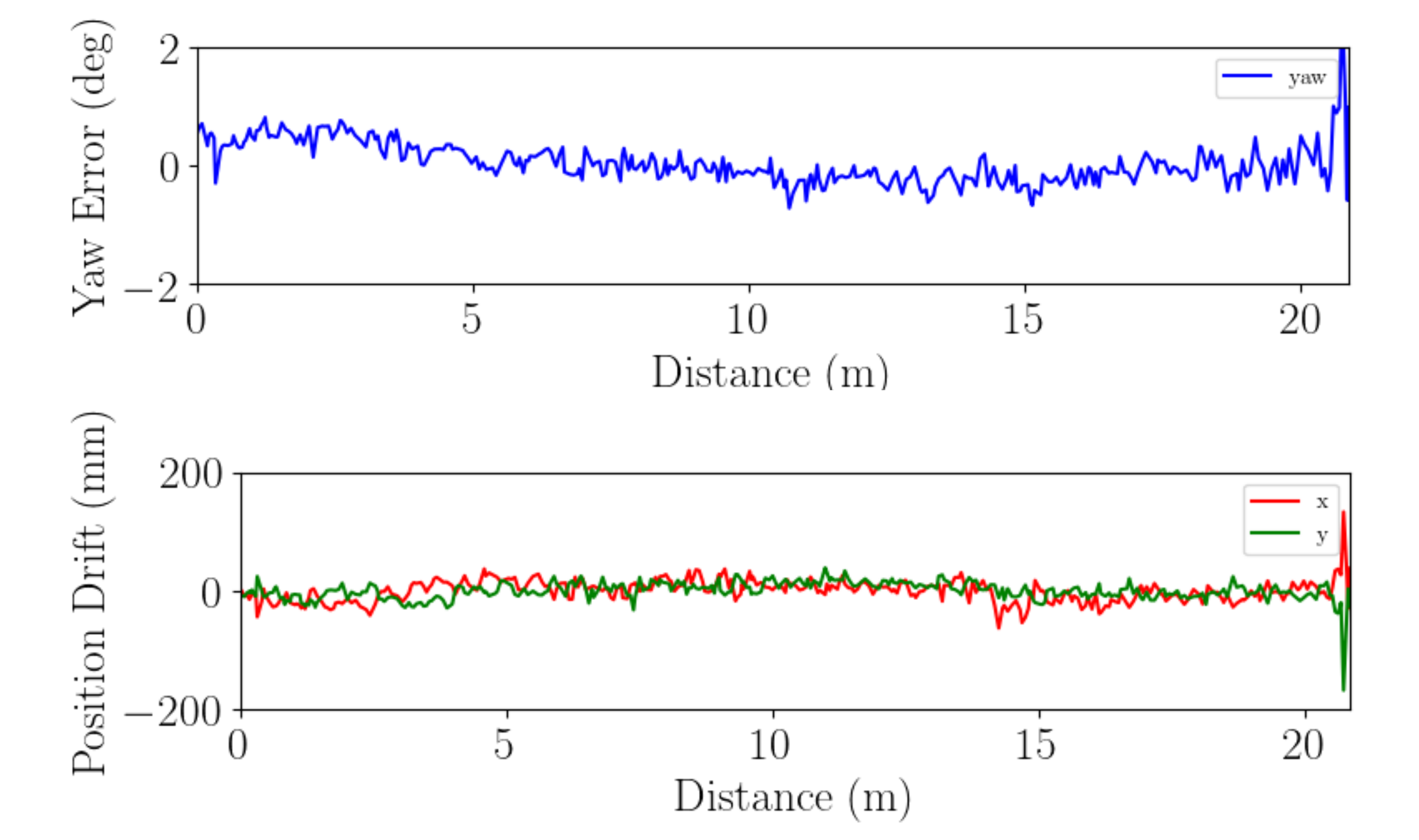}}
\caption{Error variation on Sequence 5-1. Red, green and blue lines represent errors of x, y and yaw estimation with respect to traveled distances.}
\label{traj5-1}
\end{figure*}

\begin{figure*}[t]
\centering
\subfigure[Sequence 2-2]{ \label{case2}
	\includegraphics[width=9cm]{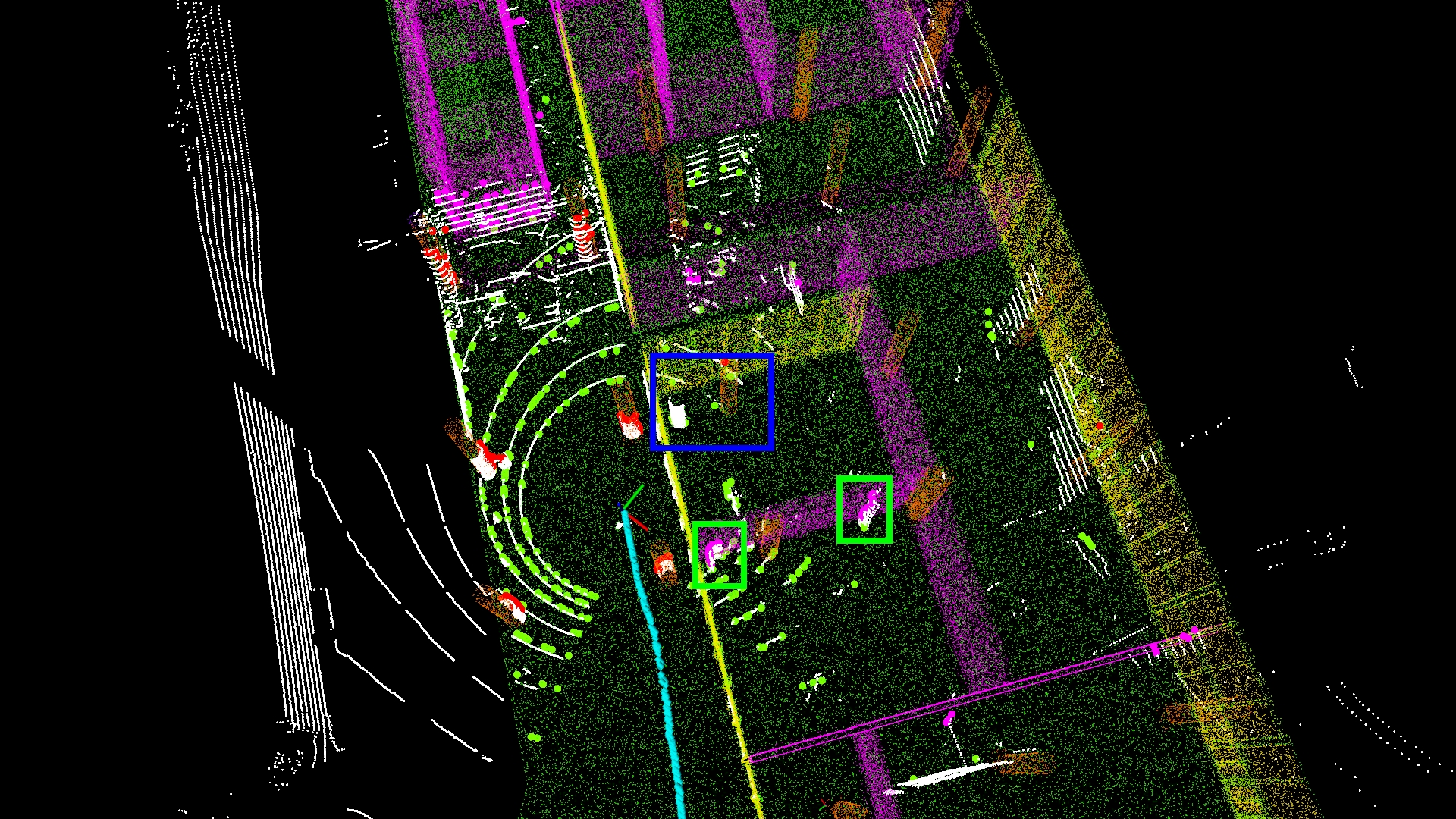}}
\subfigure[Sequence 3-3]{
	\includegraphics[width=9cm]{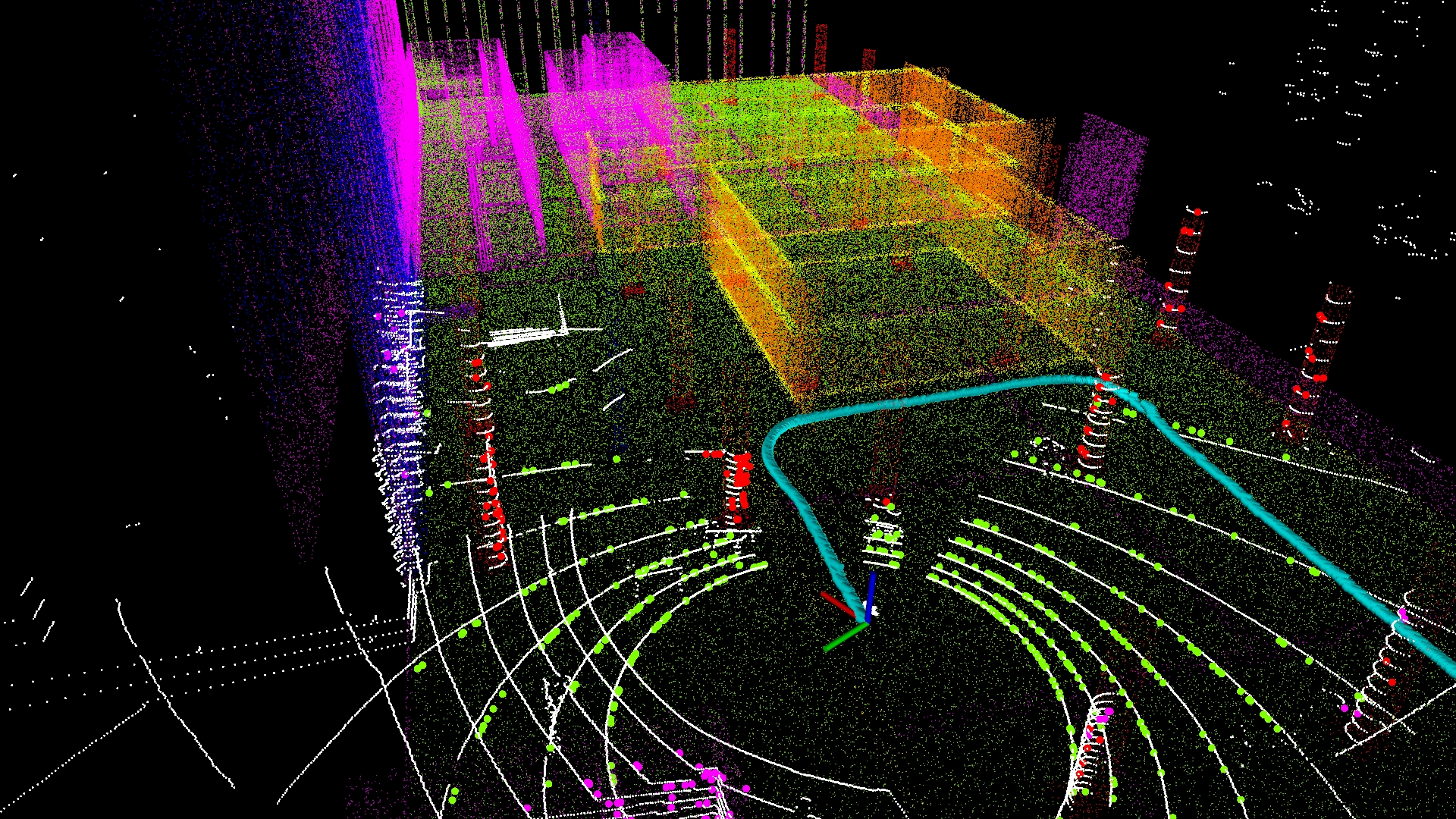}}
\subfigure[Sequence 4-1]{ \label{case4}
	\includegraphics[width=9cm]{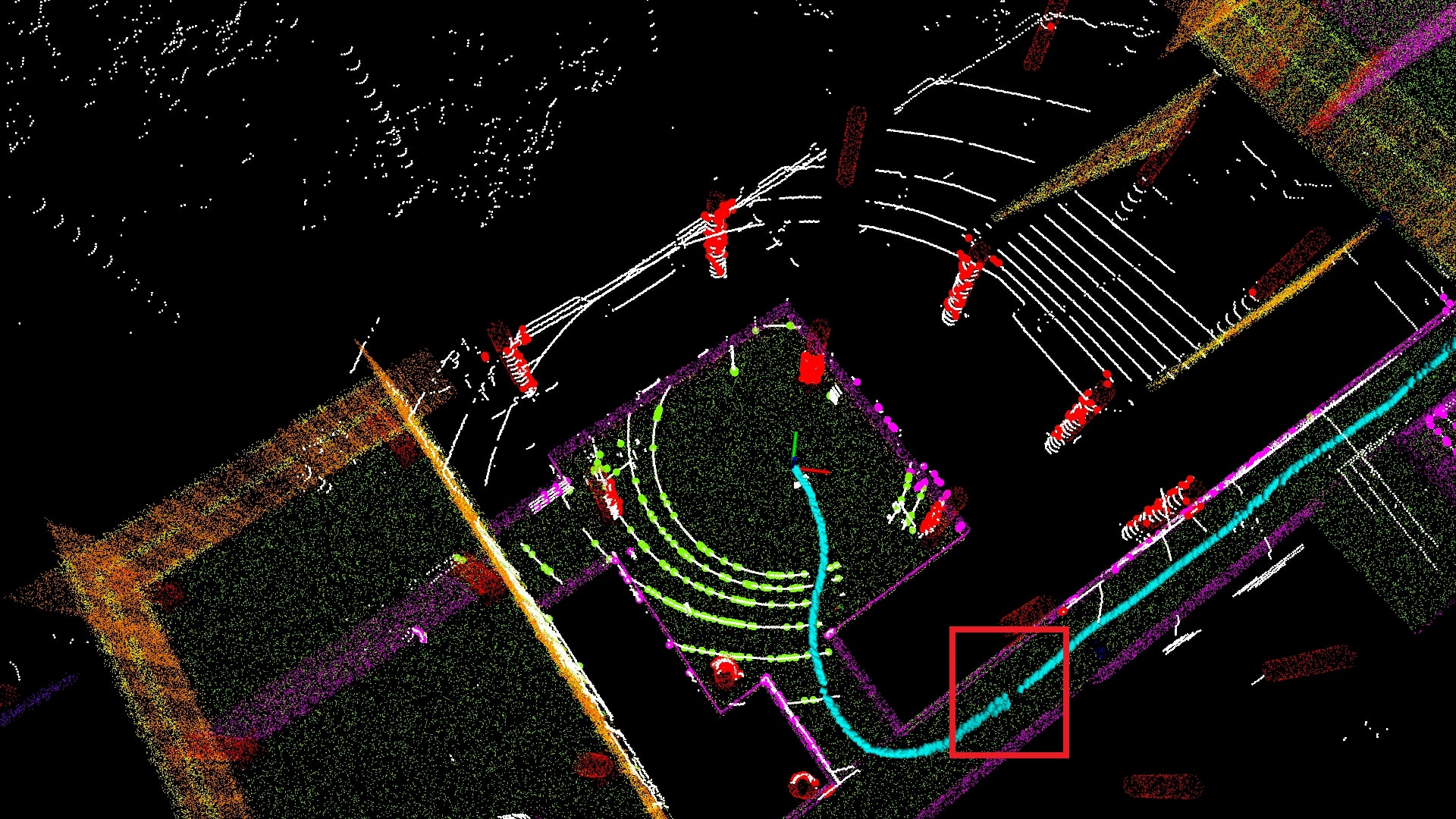}}
\subfigure[Sequence 5-1]{
	\includegraphics[width=9cm]{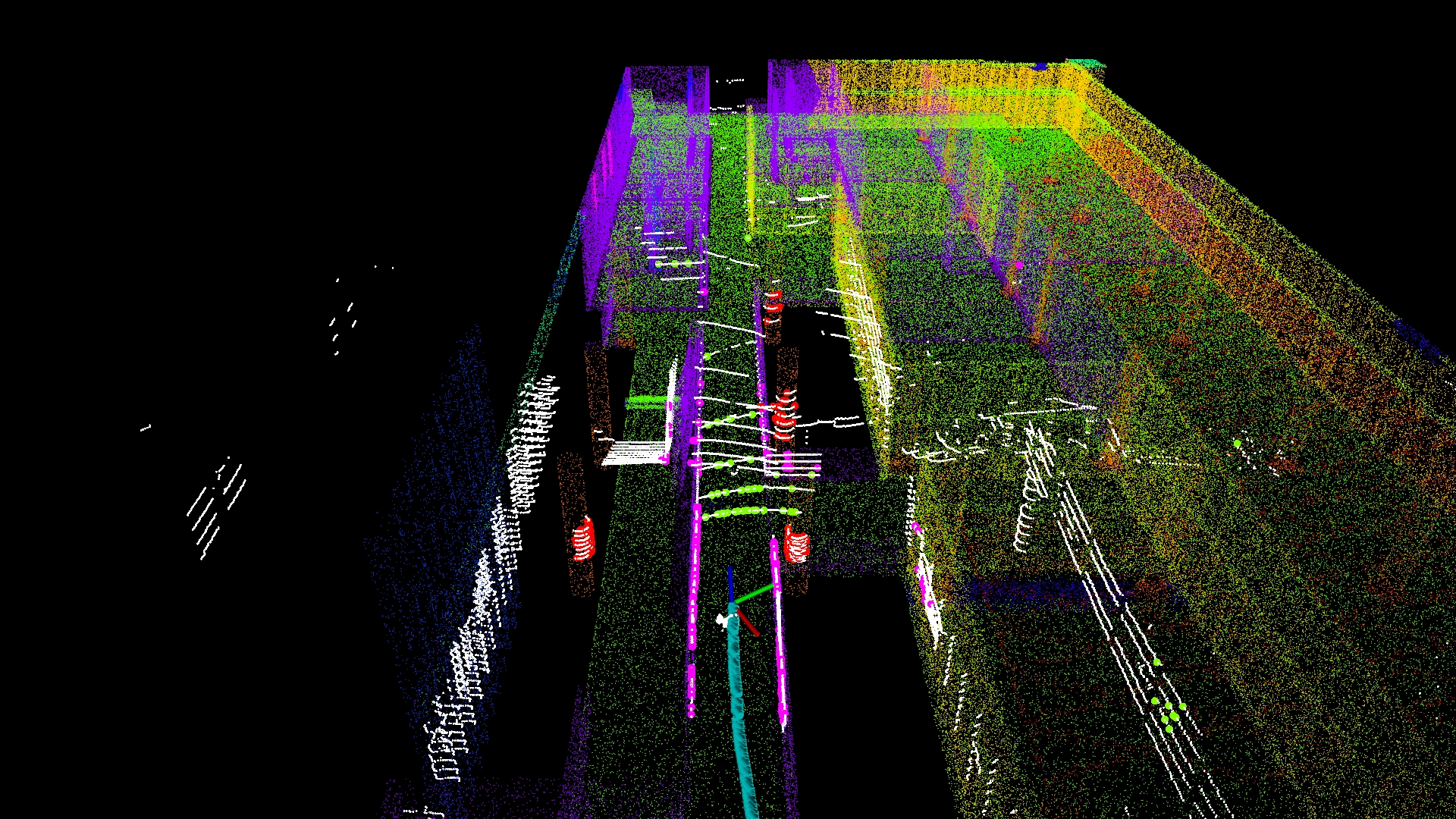}}
\caption{Screen captures in ROS Rviz. White points are input LiDAR scan $\mathcal{P}$ without data filters applied, cyan-colored trajectories are estimated poses $\mathbf{T}_{k=1,2,\cdots}$, semantic map points $\mathcal{M}$ are with various colors and matched points $\mathcal{P}^\prime$ are with larger size and colors: red for Columns, pink for Walls and green for Floors. Boxes in Figure~\ref{case2} are deviations discussed in Section~\ref{Discussion}. Red box in Figure~\ref{case4} is a discontinuity discussed in Section~\ref{Discussion}.}
\label{case}
\end{figure*}

Our proposed semantic localization pipeline consists of three steps: original ICP, semantic filtering and semantic-aided ICP with $w_cw_\rho$. We test several combinations of these steps to validate the effectiveness gradually. All the tested methods are listed as follows:
\begin{itemize}
\item ICP (ORG) \cite{besl1992method,pomerleau2013comparing}, which is actually the first step in Algorithm~\ref{semanticicp}. A common configuration in libpointmatcher \cite{libpointmatcherfilter} includes two geometric-based outlier filters: \textit{TrimmedDistOutlierFilter} and \textit{SurfaceNormalOutlierFilter}. These outlier filters are essentially weight functions. We keep this original configuration as a purely geometric-based competitive method.
\item ICP ($w_\rho$): we replace the original filters using the weight function $w_\rho$. We set $\delta=0.05$ for the test, which means the weights are with $1$ when the distances are smaller than 5cm.
\item Sem (ORG): we add the semantic labeling and selection based on the ICP (ORG). After filtering, only selected laser points are used in the second half of ICP (ORG). 
\item Sem ($w_c$), which is an updated version of Sem (ORG). In the second half of ICP, two geometric-based outlier filters are replaced by the semantic-aided weight function $w_c$. We set $\mu=0.8$ as a constant value in the test.
\item Sem ($w_\rho$): we also test the weight function $w_\rho$ under the semantic filtering scheme.
\item Sem ($w_cw_\rho$), which is the complete version of Algorithm~\ref{semanticicp} with the three steps: ICP (ORG), semantic filtering and semantic ICP.
\end{itemize}

To achieve a fair comparison, all the methods above share the same data pre-processing and filtering. Maximum number of iteration (MaxIt) is critical for ICP-based localization. For ICP (ORG), we set MaxIt as 40. As for semantic localization, MaxIt of ICP (ORG) and semantic ICP is fixed as 20 respectively, so there are also 40 iterations for a fair comparison.

Before tests on all sequences, one important configuration is to decide which elements should be used in the semantic filtering. We test the Sem (ORG) with several combinations on Sequence 3-1 in the SDE4 building, and present localization results in Table~\ref{point_selection}, in which ALL CPNT means semantic filtering is not used and all components are integrated into ICP-based localization. As observed from the table, the errors increase when windows and curtain panels are selected for localization, and decrease gradually when walls or columns are integrated into the localization. The results indicate that walls and columns could be more helpful and informative for the proposed semantic localization in the SDE4 building.

Therefore, floors, walls and columns are selected for semantic filtering based on the results and analyses above. Then, all the methods are tested on ten sequences, and errors are presented in Table~\ref{rmse}. We summarize the conclusions as follows by analyzing the results from ICP (ORG) (Column 1) to Sem ($w_cw_\rho$) (Column 6):
\begin{itemize}
	\item ICP (ORG) and ICP ($w_\rho$) results in an acceptable accuracy. This indicates that it is feasible to achieve LiDAR localization on BIM-generated maps with only one LiDAR sensor.
	\item Sem (ORG) performs better than ICP (ORG) on most sequences, indicating that the proposed semantic filtering can help improve LiDAR localization. There is an overall improvement of 18\% on the translation error.
	\item Compared to Sem (ORG), Sem ($w_c$) achieves higher accuracy on translation estimation but lower accuracy on heading estimation. This indicates that the semantic-aided weight function $w_c$ almost has the same performance as the two geometric-based outlier functions, but it could not improve the original method significantly.
	\item Compared to Sem (ORG) and Sem ($w_c$), Sem ($w_\rho$) performs the best under the semantic filtering scheme. We consider it is because the Huber function $w_\rho$ can better overcomes the deviations. 
	\item The complete version Sem ($w_cw_\rho$) could not improve the overall performance compared to Sem ($w_\rho$), but it can handle a challenging sequence 4-2 in the datasets. Finally, Sem ($w_cw_\rho$) can achieve an overall improvement of 34\% on translation estimation compared to the original version ICP (ORG).
\end{itemize}

More specifically, we can find that the localization errors of the 2nd and 4th Storey are higher than that of the 3rd and 5th. This indicates that localization difficulty is related to the accuracy of maps and environments, as analyzed in Section~\ref{data}.

The translation errors are also presented in Figure~\ref{error_boxplot} using boxplots. The localization performance can be visualized from the median error and the error variance in the boxplots. The localization errors decrease when the semantic filtering is applied on ICP (ORG), thus verifying the hypothesis that using semantic properties can improve localization. The evaluation package \cite{zhang2018tutorial} also provides variations on errors with respect to the traveled distance. The estimated localization trajectories and errors are shown in Figure~\ref{Trajectories},~\ref{traj4-1} and~\ref{traj5-1}. As observed from the trajectories and errors, Sem ($w_cw_\rho$) results in a smooth trajectory close to the ground truth, thus verifying the effectiveness of the proposed coarse-to-fine localization pipeline.

In addition to the numerical analyses, we present several case studies of localized LiDAR scan on BIM-generated semantic maps, as shown in Figure~\ref{case}. The four cases show the localization process of Sequence 2-2, 3-3, 4-1 and 5-1 in different storeys of the SDE4 building. We also present the number of points in each step of these four cases, shown in Table~\ref{points}. After the random sampling and semantic filter, only hundreds of laser points ($\approx$3\% of raw data) are selected for the final semantic ICP step. As for efficiency, the mean-time cost of semantic localization (Algorithm~\ref{semanticicp}) is 108ms, 79ms, 112ms and 114ms in these four sequences. The real-time method is able to track the LiDAR scanner operating at 10Hz with only a resource-constrained embedded device. We also release a video demonstration online \footnote{The video is available at \href{https://youtu.be/6jscy2Y5mj4}{this link}}.

\begin{table}[!t]
\captionsetup{justification=centering}
\renewcommand\arraystretch{1.5}
\begin{center}
	\caption{Number of Laser Points in Figure~\ref{case}}
	\label{points}
	\begin{tabular}{p{1.0cm}<{\centering}|p{1.8cm}<{\centering}p{1.8cm}<{\centering}p{1.8cm}<{\centering}}
		\hline
		\hline
		Case & raw $\mathcal{P}$ & datafilter$\left(\mathcal{P}\right)$ & $\mathcal{P}^\prime$ after (\ref{secondstep})\\
		\hline
		2-2 & 23375 & 2099 & 679 \\
		3-3 & 22932 & 2204 & 897 \\
		4-1 & 21209 & 1910 & 660 \\
		5-1 & 25428 & 1569 & 487 \\
		\hline
		\hline
	\end{tabular}
\end{center}
\end{table}

\subsection{Compared to LiDAR-only SLAM systems}
\label{compare}

\begin{table*}[t]
\captionsetup{justification=centering}
\renewcommand\arraystretch{1.5}
	\begin{center}
		\caption{Localization accuracy using SLAM and BIM-based localization}
		\label{slam}
		\begin{tabular}{p{0.8cm}<{\centering}|p{0.75cm}<{\centering}p{0.75cm}<{\centering}p{1.1cm}<{\centering}|p{0.75cm}<{\centering}p{0.75cm}<{\centering}p{1.1cm}<{\centering}|p{0.75cm}<{\centering}p{0.75cm}<{\centering}p{1.1cm}<{\centering}|p{0.75cm}<{\centering}p{0.75cm}<{\centering}p{1.1cm}<{\centering}}
			\hline
			\hline
			\multirow{2}*{Seq.} & \multicolumn{3}{c|}{LOAM \cite{zhang2014loam}} & \multicolumn{3}{c|}{DLO \cite{chen2022direct}} & \multicolumn{3}{c|}{Open3D SLAM \cite{jelavic2022open3d}}  & \multicolumn{3}{c}{BIM-based Localization}\\
			&Tr.(m) &Rt.($^\circ$) & $\Delta$ Z(m) &Tr.(m) &Rt.($^\circ$) & $\Delta$ Z(m) &Tr.(m) &Rt.($^\circ$) & $\Delta$ Z(m) &Tr.(m) &Rt.($^\circ$) & $\Delta$ Z(m) \\
			\hline
			2-3  & \textbf{0.058}  & 0.296 & -0.968 & 0.062 & \textbf{0.263} & -2.675 & 0.100 & 0.280 & -1.547 &0.077 &0.590 & \textbf{0.084}\\
			3-3 & 0.040 & \textbf{0.302} & -0.010 & 0.034 & 0.396 & -0.025 & 0.046 & 0.662 & 0.015 & \textbf{0.030} & 0.359 & \textbf{-0.002}\\
			4-2  & 0.038 & 0.326 & -0.793 & \textbf{0.036} & 0.291 & -0.981 &0.041 & \textbf{0.276} & -0.874 &0.079 & 1.198 & \textbf{-0.052}\\
			5-2  & \textbf{0.017} & 0.303 & -0.938 & \textbf{0.017} & \textbf{0.182} & -1.149 & 0.033 & 0.215 & -0.709 &0.036 & 0.345 & \textbf{0.083}\\
			\hline
			\hline
		\end{tabular}
	\end{center}
\end{table*}

\begin{figure*}[!t]
\centering
\subfigure[Top view]{
	\includegraphics[width=18cm]{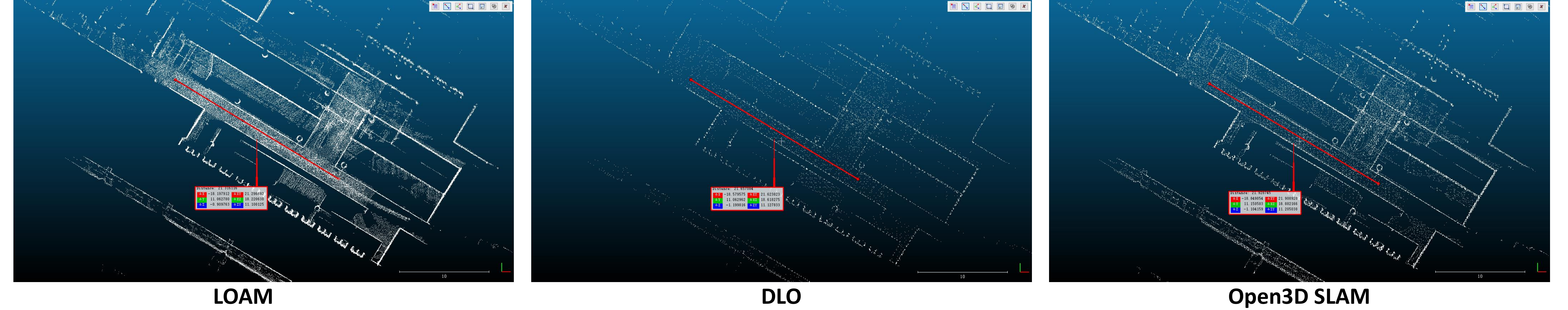}} \\
\subfigure[Side view]{
	\includegraphics[width=18cm]{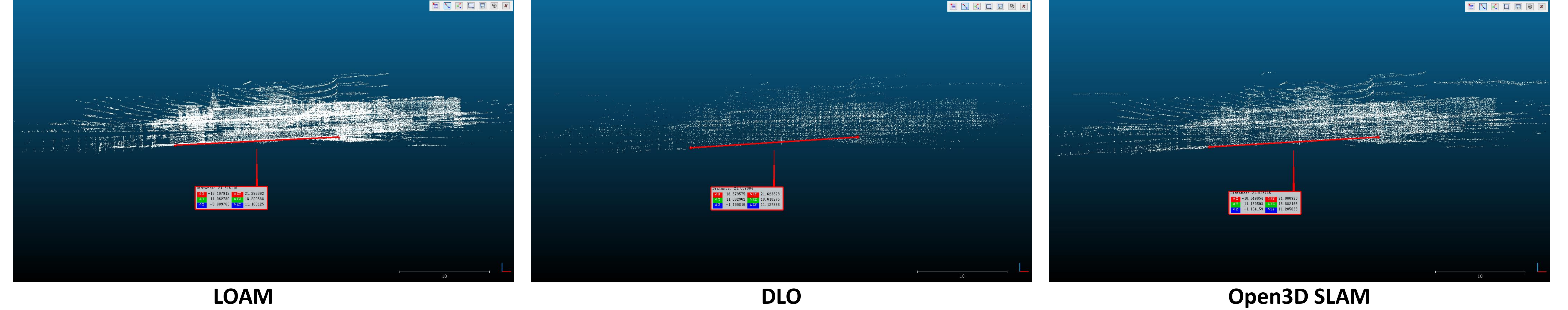}}
\caption{Maps generated from LOAM, DLO and Open3D SLAM using Sequence 5-1 data. Red lines are used for visually displaying drifts on Z-axis.}
\label{slam-5-1}
\end{figure*}

SLAM is a widely-used mapping system that aims to achieve localization and mapping simultaneously. In this subsection, we also compare the proposed BIM-based localization with the SLAM systems. Three open-sourced SLAM systems are performed on our self-collected data: Lidar Odometry and Mapping (LOAM) \cite{zhang2014loam} \footnote{https://github.com/HKUST-Aerial-Robotics/A-LOAM}, Direct Lidar Odometry (DLO) \cite{chen2022direct} \footnote{https://github.com/vectr-ucla/direct\_lidar\_odometry} and Open3D SLAM \cite{jelavic2022open3d} \footnote{https://github.com/leggedrobotics/open3d\_slam}. Specifically, LOAM and DLO are LiDAR-based odometry methods (reduced LiDAR SLAM systems), while Open3D SLAM is a complete LiDAR SLAM system, including loop closing and graph optimization. For a fair comparison, there is no IMU or other information as assistance used in this experiment.

The SLAM systems are first evaluated by comparing them to the ground truth poses in 2D space, and the translation and orientation errors can be obtained accordingly. These errors are eventually calculated on 2D x-y plane using Equation~\ref{rmse1} and \ref{rmse2}. In addition, we also propose to evaluate the drifts on the Z-axis of SLAM systems and our proposed BIM-based localization. The drift errors are calculated under the criteria: $\Delta Z =  Z_{last} - Z_{init} $ (m), in which $Z_{init}$ and $Z_{last}$ are the average height of 50 poses at the beginning and the end of the trajectory. The data collection process is conducted with a planar motion. Thus $\Delta Z$ can be regarded as a measurement metric to evaluate the drift on Z-axis.

The quantitative results on four sequences are presented in Table~\ref{slam}. Compared with the LiDAR-only SLAM systems, the proposed BIM-based lidar localization does not show better performance on 2D pose estimation. Still, it shows competitive results by matching scans on BIM-generated maps. On the other hand, LiDAR-only SLAM systems show more significant height drifts in Sequence 2-3, 4-2 and 5-2, because these trajectories contain few revisted places for loop closing, as shown in Figure~\ref{sequence_traj}. The drift on Z-axis is a common degeneracy problem for LiDAR-only SLAM applications, especially in a long straight travel. We also present the SLAM-generated maps in Figure~\ref{slam-5-1}. While in the BIM-generated maps, the floor is almost perfectly flat in one storey, which means all floor points are with the same height, thus providing certain constraints for BIM-based localization.

\subsection{Discussion}
\label{Discussion}

In Figure~\ref{case2}, it is interestingly found that there are notable differences between pre-built $\mathcal{M}$ and observed $\mathcal{P}$, which are essentially the differences between as-designed and the as-built. The two columns in green boxes are observed in the LiDAR scan but there are no columns on the map respectively, making the LiDAR points match to walls (colored with pink) due to the nearest neighbor search strategy of ICP. Another observed column in the blue box is not matched to any element since there is a considerable distance between the nearest column on the map. Actually, the mismatch problem occurs on every storey in the NUS SDE4 building because of the deviations between as-built and as-designed. There are other factors that cause errors in this study, such as dynamics and sensor noises. On the other hand, the localization pipeline is designed with powerful weight functions, so it can still track the pose successfully under these challenges.

We also notice that there is a large error when traveling in the long, challenging corridor on the 4th Storey. This results in a discontinuity in the estimated trajectory, shown in the red box of Figure~\ref{case4}. Specifically, the large drift is not eliminated in a short time. There are mainly two reasons for this. Generally, a long corridor is a challenging scene for localization that will degenerate the localization performance. Besides, once a pose is with a large error, the considerable error may be conducted into the following pose estimation.

Overall, there are still some requirements and challenges when applying the proposed localization method on BIM-generated maps. We first summarize the requirements for applying our proposed framework:

\begin{enumerate}[label=(\alph*)]
	\item The proposed method is applicable in static built environments. In a dynamic environment, like an ever-changing construction site, the BIM model should be reviewed and updated by the user, which will involve human labor and be time-consuming in application. 
	\item The BIM content should contain the basic geometric sizes and category labels of main structures in a building. These two pieces of information are necessary requirements to generate semantic point cloud maps in this study.
	\item It is unavoidable that there exist deviations between as-designed and as-built. The deviations should not be too large in the application for localization success.
	\item The proposed method is more appropriate to use in environments that have certain diversity. This diversity includes the categories and spatial distribution of BIM elements. For example, a typical scene is a long straight corridor which consists of only walls and floors, which is lack of diversity and is challenging for LiDAR-only localization.
\end{enumerate}

We also list the challenges and limitations of the proposed method as follows:

\begin{enumerate}[label=(\alph*)]
	\item The biggest challenge is the deviations between as-designed and as-built. The deviations can cause incorrect data associations and ambiguous scans, which could cause localization failure in challenging scenes.
	\item There exist inaccurate semantic maps using the proposed BIM-to-Map conversion, as shown in Figure~\ref{error_boxplot}. These incorrectly or mixed labeled map points may reduce the diversity of the semantics, leading to a degeneration of localization accuracy. 
	\item The pose estimation of $\mathbf{T}_{k}$ relies on the result of $\mathbf{T}_{k-1}$, which means the significant error in $\mathbf{T}_{k-1}$ may also result in $\mathbf{T}_{k}$, or even cause a localization failure. In addition, we manually set the floor and $\mathbf{T}_{k=0}$ at the first stamp of each sequence. To build a more automatic localization system, we need to estimate the initial pose in a whole building with a global localization module \cite{yin20193d,dreher2021global}.
\end{enumerate}

\section{Conclusions}
\label{conclusion}

This paper proposes a mapping-free and learning-free semantic localization framework. A BIM-to-Map conversion is proposed by using spatial locations and category labels of elements in BIM. This paper also proposes a coarse-to-fine localization method to track a 3D LiDAR sensor based on semantic maps, in which both geometric and semantic information are considered in data associations. The tests on real-world datasets demonstrate that the proposed framework can achieve effective and efficient localization using only one BIM file and one mobile LiDAR sensor.

We consider there remain research directions based on the experimental results and discussion on limitations in Section~\ref{Discussion}. We list some of them as follows:

\begin{itemize}
\item The accuracy of semantic map building and data labeling can be improved by integrating the geometrics of local point clouds, e.g., the points with similar surface normals might be in the same category.
\item Another promising study is that we can first filter certain elements in BIM first \cite{follini2020bim}, and then generate semantic maps. It is also worth studying how to filter the BIM to guarantee the localization performance.
\item Multiple sensors can help improve the robustness and accuracy of localization, e.g., IMU as an assistance and support to overcome the Limitation (c) and (d).
\item To address the Limitation (e), global registration or localization is critical for applications, which can localize a robot from scratch without initial guess.
\item Besides, the map management is also important in large indoor scenes. A concise and interactive map form is desired for robot navigation, such as topological maps.
\end{itemize}

\section{Acknowledgement}
\label{ack}

This research is supported by Building Construction Authority (BCA) and National Robotics Programme (NRP) under its Built Environment Robotics R\&D programme (Grant Award Ref No. W2122d0154). Any opinions, findings and conclusions or recommendations expressed in this material are those of the authors and do not reflect the views of the BCA and NRP.

This project is also supported by the Hong Kong Center for Construction Robotics (InnoHK center supported by Hong Kong ITC).

We would also like to thank colleagues in NUS ARC for kindly sharing the experimental devices with us.

\bibliographystyle{elsarticle-num-names}
\bibliography{root}


\end{document}